# An Algorithm for Repairing Low-Quality Video Enhancement Techniques Based on Trained Filter

**Student Name: Lijun Wang**
**Supervisor: Dr. Ling SHAO**

**Date: 31/08/2010**

**This dissertation is a part requirement for the degree of M.Sc. in Data Communication**

# ABSTRACT

Multifarious image enhancement algorithms have been used in different applications specifically. Still, some algorithm and modules are imperfect for practical use. When the image enhancement modules have been fixed or combined by a series of algorithm, we need to repair it as a whole part without changing the inside. This report aim to find an algorithm based on trained filters to repair low-quality image enhancement modules. A brief review on basic image enhancement techniques and pixel classification methods will be presented, and the procedure of trained filter will be described step by step. The experiments and results comparisons for this algorithm will be described in detail.

# ACKNOWLEDGEMENTS

My deepest gratitude goes first and foremost to my supervisor, Dr. Ling Shao, for his constant encouragement and guidance. Under his help, I am getting interest increasing in this subject, and would like to learn more about digital image processing algorithms. This is very important for me because I believe the interest is the motive power of research.

I also want to thanks my friends and my fellow classmates who sharing time with me on discussing the project problems and always give me great encouragement during the duration of this project.

# CONTENT

# FIGURE CONTENT

**Department of Electronic & Electrical Engineering**

MSc Project

DECLARATION

(To be completed by the student and inserted into the dissertation immediately after the contents page)

I certify that all sentences, passages and figures / diagrams quoted in this thesis from other people's work have been specifically acknowledged by clear cross-referencing to the author(s), work and page(s).

Furthermore I have read and understood the definition of Unfair Means for assessed work produced in the M.Sc. Electronic Engineering Student Handbook (page 18/19) and have complied with its requirements.

I understand that failure to comply with the above amounts to plagiarism and will be considered grounds for failure in this thesis and the degree examination as a whole.

Name (please use blocks capitals)    WANG LIJUN

Signature

Date    31$^{st}$ August

# Chapter 1  Introduction

Image and video processing has been developed rapidly as an important research field at present, since demanded by various and numerous areas of applications such as in biology, archaeology, medicine, spaceflight, and display industry. Images and video enhancement is one of the most important and interesting area of video processing. We already have numerous processing algorithms and modules to enhance images. For all that, these algorithms or modules are usually imperfect. Exceptionable results could be produced, if they didn't tune appropriately. In practical, the processing module often has been fixed or work as an integrated algorithm. We need to repair them on the ground of the exits processing modules.

To address this problem, we optimally repair the images created by the previous video processing modules based on Least Mean Square filters. Recently, Classification-based least squares (LS) filters, called "trained filters" [1], have given impressive performances for image enhancement applications including resolution up-scaling and coding artifacts reduction. To repair the output images, firstly, we can use LMS optimization algorithm to obtain corresponding optimized filter coefficients, then simulate the previous video enhancement processing, and repair the processed image by using the filter coefficients obtained before. In addition, for particular processing modules, we should design an appropriate classification method individually to get a better performance.

The aim of this project is to carry out three embodiments of repairing image enhancement algorithms based on trained filters. These applications are coding artifacts reduction, resolution up-scaling, and deblurring. The corresponding algorithms and classification methods are selected separately, and achieved by programming in C language. An image quality assessment named SSIM was carried out as a program which can ready 4:2:2 YUV sequence in MATLAB, At last, the comparison of the input and out sequences with an evaluation will be given to show the repaired level.

This report will summarize some fundamental knowledge of digital image processing used in this project in Chapter 2, which is including basic algorithms in solving image enhancement, classification methods, image quality assessment, and process of trained filter. All of these will be used in simulating the proposed algorithm in programming. In Chapter 3, As the Core Concepts, the framework of proposed algorithm will be expressed step by step to help getting a clear vision for the procedure of the experiment. Moreover, the parameters settings and method selecting during simulation will be described vivid. The LS optimization algorithm used in the training process will be expressed as well. Chapter 4 will present the experiment results and evaluation results. A comparison of the input and output from different processes will be given in each embodiment. Conclusions and further work will be summarized in Chapter 5. For a conciseness reference, the outputs of repairing tests sequence will be attached at the end.

# Chapter 2  Lecture Review

Digital Image Processing was first required in the newspaper industry and space project in the early 20th century. This discipline has grown rapidly, which based on the development of digital computers and the highly increased backing techniques including: data store, display, and transmission. It has been used in a boarder range of applications whereas in the past, such as blurred pictures were sharpened by image processing method in archeology; degraded images were repaired by image enhancement procedures in geography [2].

Digital Image Processing consists of the many aspects, such as image enhancement, image restoration, image acquisition, compression etc [2]. This project will focus on researching the image enhancement and image restoration. The relevant processing algorithms and also the quality assessments to evaluate them is represented in this chapter.

## 2.1   Image enhancement and restoration

Image enhancement and image restoration both aim to get image improvement. Image enhancement techniques [3] are aiming to modify attributes of an image and make it more suitable for different applications while Image restoration techniques focus on restoring an image degraded by blurring, noise or coding artifacts, down-sampling, geometry distortion, etc. A large number of image processing approaches are available for the two aspects, and the algorithms are usually designed for specific applications.

Three main applications belonging to the two areas will be discussed in this project. They are coding artifact reduction, resolution up-scaling, and deblurring. For a flagrant contrast between the output and input, we will choose those basic and low-cost algorithms for experiment. Advanced algorithms are just for reference. A briefly summary of that is in the following.

### 2.1.1   Coding artifacts reduction

In spatial domain, soothing linear filters such as averaging filter [4] and smoothing nonlinear filters including median filter is employed. For example, averaging filters is based on reassign each pixel location with the average value of its neighbors in a preset filter aperture. In this way, sharp transitions in gray levels (or intensity) are reduced. Figure 1 is showing a result of using median filter to reduce coding artifacts.

In frequency domain, sharp transitions in intensity, such as block, noise or coding artifacts, usually present as high frequency content in Fourier transform. Therefore Low pass filters can be used as smoothing filters to reduce them. Three types of low pass filters were considered to solve this problem. They are idea, Gaussian and Butterworth [2].

However, desired detail, such as edges, may be smoothed at the same time. To get better results, edge-adaptive method should be added.

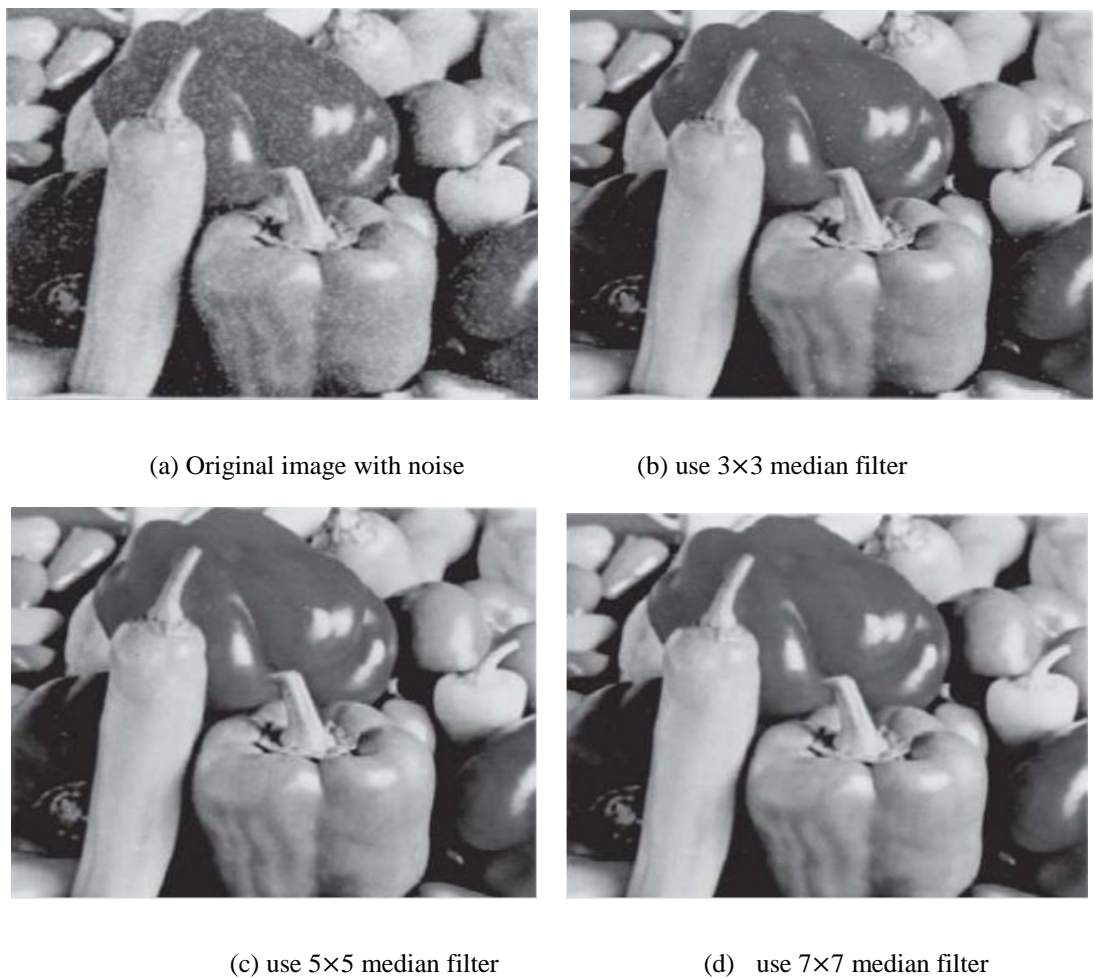

<div align="center">

(a) Original image with noise     (b) use 3×3 median filter

(c) use 5×5 median filter     (d)  use 7×7 median filter

Figure 1   Using median filter to reduce noise [5]

</div>

### 2.1.2   Deblurring

Deblurring is also known as sharpness enhancement. Sharpening is the opposite operation of blurring.

In spatial domain, while for blurring, we use averaging method, we could use mathematical models to do differentiation for sharpening. Generally, detail-sharpening spatial filters are based on first derivative or second derivatives, such as the Gradient (using the first derivative) and the Laplacian (using the second derivative) [2].

In frequency domain, high pass filter was implemented to weak low frequency content and protect high frequency content at the same time in Fourier transform. Four types of that are mainly used. They are idea, Butterworth, Gaussian, and Laplacian [2].

Some other advanced method such as un-sharp masking and high-boost filtering are used widely in practical. Figure 2 is showing a result by using un-sharp masking method.

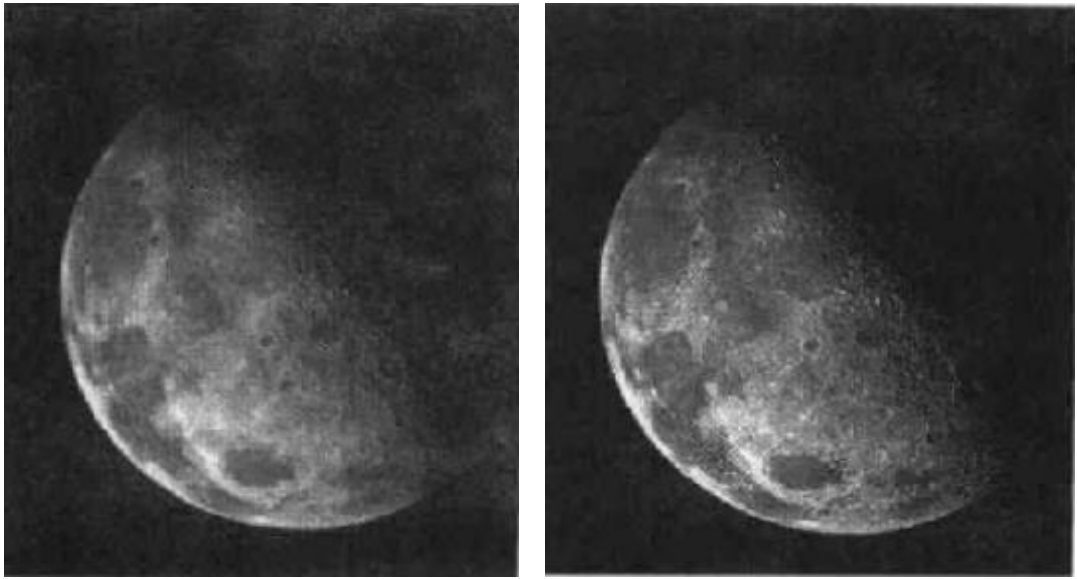

Figure 2    The right is original image, and the left is sharpened image. [2]

### 2.1.3    Resolution up-scaling

Resolution up-scaling or resolution up conversion is converting an image with low resolution into higher resolution.

In spatial domain, basis method for this application is increasing the number of the pixels by using image interpolation which we will mention in subsection 2. The most of the generalized interpolation techniques [6] are linear resolution up scaling algorithm. Figure 3 is showing a result of resolution up-scaling by interpolation technique.

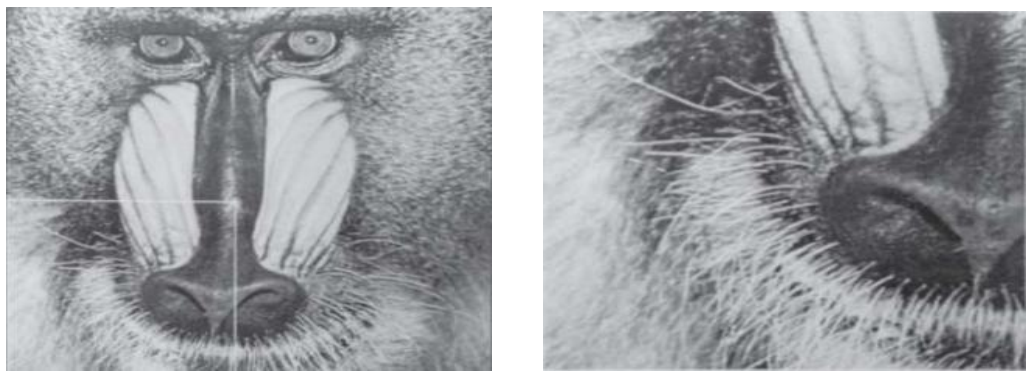

The right image is the original one. The left one used interpolation method for 2:1 magnification. [5]

Figure 3    Interpolation for 2:1 magnification

In frequency domain, the effect of resolution up-conversion is similar with enhancing the sharpness after increasing pixel locations. So we can also employ high pass filter to hold the high frequency content and get a better performance.

Recently, the advanced non-linear resolution up-scaling algorithm developed rapidly, such as Kondo's data dependent interpolation filter [7], CBA [8].

## 2.2 Least Mean Square Optimization

### 2.2.1    Solving Least Squares Problem

Least Square Algorithm is "the standard approach has been to use partial derivatives to obtain necessary conditions for a minimum, and then solving the equations involved" [10]. This Algorithm is most used in data fitting problem including extrapolation, smoothing of data, and approximate model of data.

We just focus on the approaches to solve the matrixes here which will be used in the follow subsection. There are four approaches mainly used to solve this problem. They are Gaussian elimination, QR transformation, singular value decomposition (SVD), and conjugate gradient. The Gaussian elimination algorithm may be quicker than QR transformation if the matrix is not square. But QR which used orthogonal decomposition is more stable. SVD can give indication how independent columns are. Conjugate gradient is better for large sparse problem.

### 2.2.2    Least Mean Squares Filter

LMS filter aim at minimizing the mean square error (MSE) between the target image and the ideal image. Among numerous video enhancement algorithms, Least Mean Square Filtering techniques give out impressive experimental results, including: Kondo's classification-based LMS [11] filter algorithm and Li's localized LMS filter algorithm [12] on resolution up-conversion applications. Least Mean Square Optimization method which has strong theoretical background has an important advantage that the parameters used here usually can be calculated from a given degraded image [2].

## 2.3 Trained Filter

Trained Filter is a low quality image enhancement repairing algorithm. The flash point of this algorithm is using classification-based least square algorithm to obtain optimization coefficients which is used in repairing procedure. The coefficients used here are obtained from training "on the combination of target image and the degraded version thereof that act as the source" [1]. "The trained filter consists of two parts are the off-line training process and the run-time filtering process" [9].

### 2.3.1    The off-line training process

The off-line training process is depicted in Figure 5. In this process, the degradation usually has to be individually designed as the inverse of the video enhancement processing module. For example, if the video processing module is coding artifacts reduction, the degradation in this embodiment should be compression or codec. After this, the image degraded from the target image will be processed by using the previous video processing module which needs to be repaired. The following operation is classification which need be selected or designed according to the specific video enhancement algorithm. The selection of classification method for different applications will be discussed in section 5 in detail.

The pixels in degraded image will be classified, and the corresponding pixels in the target image will be accumulated. From these two data sources, the optimized coefficients can be calculated by using Least Square minimization. The Least Mean Square Optimization used in this training process will be described in section 4. After the training process, we will obtain the optimized coefficients been stored for each class in look-up table (LUT) for repairing.

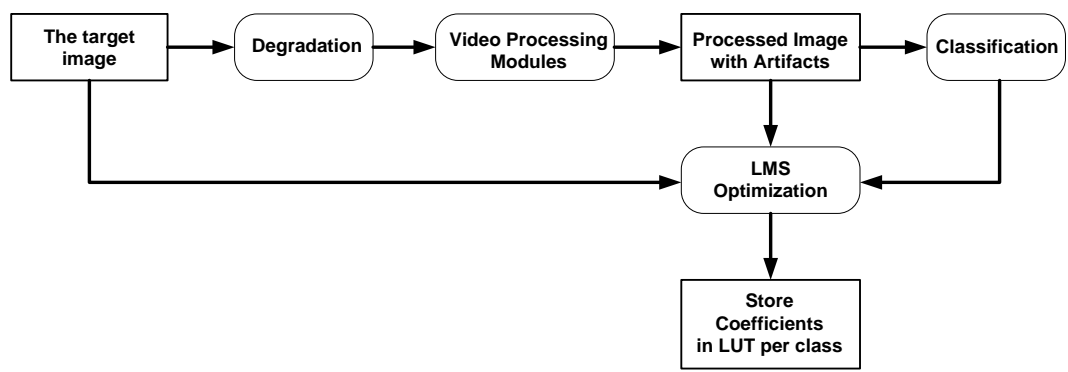

Figure 4    The off-line training process in trained filter algorithm

### 2.3.2    The run-time filtering process

The run-time filtering process is depicted in Figure 6. In this process, original low-quality image will be processed by previous image enhancement processing modules. The processed image will be filtered using the optimized coefficients after classified. The classification method used here is the same one as it is used in training process. By using the classification attained before to index the LUT, the corresponding optimized coefficients can be easily retrieved for using.

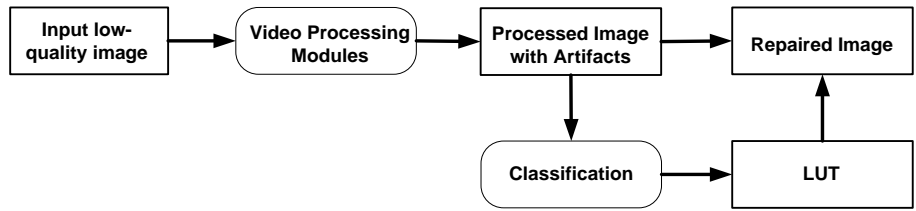

Figure 5   The run-time filtering process in trained filter algorithm

## 2.4 Pixels Classification Method

Various classification methods are used to classify the pixels in a pictured according to the various pixel properties, such as the intensity (gray-level), color, or local image characteristics calculated on the pixels' neighborhoods [13]. In this way, the differences between each class are valuably much more than the disparity in a particular class. Therefore, pix operations become effective and valuable based on classifications.

In this project, classification is used to distinguish local structures and local complexity which are two important properties used in image enhancement algorithm. A review of classification methods on basis of them will be represented in this section. At the end of this section, appropriate classification methods designed for each embodiment will be proposed.

### 2.4.1   Classification on basis of local structures

For image quality improvement, Adaptive Dynamic Range Coding (ADRC) [14] which is proposed by T. Kondo and K. Kawaguchi turned out to implement simply and possess high efficiency for representing the local structure.

In practical, brightness which is ranged from 0 to 255 as a value of specific pixel can be calculated for classification. Firstly, images are prepared by division of blocks in a fixed size. When we are encoding each pixel in the image into 1-bit with ADRC, the ADRC code of each pixel $Q_i$ in an observation block is simply defined as: If $V_i \leq V_{av}$, then $Q_i = 0$;If $V_i > V_{av}$, $Q_i = 1$. Where i indicate the index of pixel in the block; $V_{av}$ express the average value of all the pixel values, $V_i$, in the block. A $3 \times 3$ block encoded according to ADRC was shown in figure 7 as an example. In this project, the division pattern will be according to a diamond shaped filter aperture which we will discuss later.

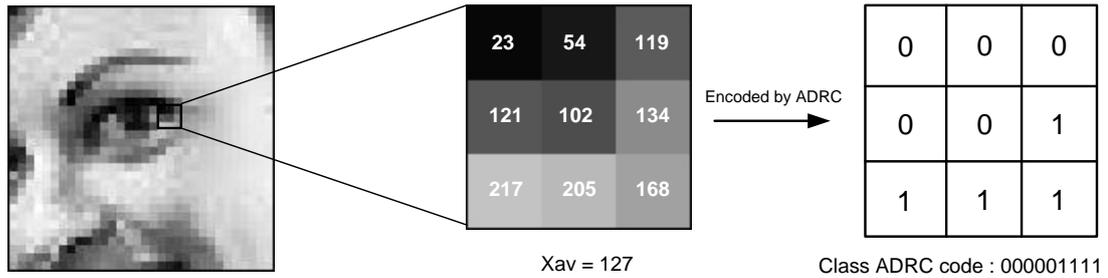

Figure 6   ADRC code of a $3 \times 3$ block

### 2.4.2   Classification on basis of local complexity

ADRC is implemented successfully for classification on basis of image structures, but some structure of noise or coding artifacts may have the same ADRC code pattern as one of the other details'. Therefore, some other classification methods are proposed to be combined with ADRC which is not enough to be single used. These classification methods including dynamic rang (DR) [15], Local entropy approach [16], Mean Absolute Difference (MAG) [17], and Standard Deviation (STD) [18]. Most of them are used for determining the complexity on a local image.

Hu and Haan [15] proposed adding the dynamic range (DR) to ADRC to solve this problem. DR is working as a preset threshold factor to determine the contrast information. Thus, one extra bit, DR, was added to the ADRC code in each block. The extra bit is defined as [15]: If $(X_{max} - X_{min}) < T_r$, then DR=0; Otherwise DR=1. Where $X_{max}$ is the maximum pixel value, while $X_{min}$ the minimum value; $T_r$ indicates a threshold value; r is a local region inside.

Unlike Hu and Haan's, Shao designed a local entropy classification approach which based on information theory to further differentiate the flat area and complex area. By using entropy equation, the local entropy H in region R can be defined as [15]:

$$H = -\sum_{i=1}^{N} P_R(i) \log_2 P_R(i) \qquad\qquad 2.1$$

Where i is the aperture index; $P_R(i)$ represents the probability of pixels having specific value in the range if the aperture i; N indicates the number of apertures.

The other two classification methods, MAG and STD, to determining the complexity are also proposed by shao [17][18]. They both turned out having obvious improvement in classification results.

### 2.5 Evaluation Method

#### 2.5.1   Image quality assessment

The operations on digital images, such as acquisition, compression, treatment, transmission and reconstruction, lead to various distortions which affect the quality of images. In this way, the quality of these images reflects the performance of the used algorithms or equipments. To optimize these systems and achieve a better performance, it is highly required to find suitable measurement for the quality of the images from both input and output.

Generally, Image quality assessment can be divided into two parts: subjective method and objective method. Since the ultimate viewer is human beings, the subjective method is more dependable, compared with objective method. However, subjective method is time-consuming and hard-sledding, and it is difficult to be applied in practice because that the different judgments will be given from different viewers. Therefore，for comparing different resulted image and dynamic monitoring them in digital image processing, objective methods are widely used by calculating image quality matrix.

### 2.5.2   MSE and PSNR

In mathematic statistics, MSE (Mean Square Error) of an estimator is a way to quantify the difference between an estimated value $\theta$ and the true value $T$.

$$MSE\ (\ T\ ) = E\ ((T - \theta)^2) \qquad\qquad 2.2$$

MSE is defined as the respect of $\theta$ to the estimated parameter T. To estimate the quality of an image, the estimator can be supposed as a sum of an original signal and a distorted error signal. In this way, the difference of two images can be calculated and quantifies the strength of the error signal.

$$MSE = \frac{1}{MN}\sum_{i=1}^{M}\sum_{j=1}^{N}[a(i,j) - \hat{a}(i,j)]^2 \qquad\qquad 2.3$$

In this formula, $a(i,j)$ and $\hat{a}(i,j)$ represent the intensity or color value of the reference image and reconstructed image; MN expresses the total number of an $M \times N$ image pixels; $a_{max} = 2^l - 1$, l which is defined as color dept usually equals to 8, $0 < a < 255$.

By calculating the average value of the squared intensity differences of distorted and reference image pixels, MSE score can simply implemented and reveal the image quality to a certain extent.

However, two images which are processed in different way may get the same MSE score. Obviously, they have very different types of errors. Also, the one which get higher MSE score may perform better according to viewer's visibility. In this situation, it is not well matched to perceived visual quality.

Another traditional method is Peak signal-to-noise ratio (PSNR) which is widely used in analogue systems by calculating a full reference quality metric. PSNR is easily defined via MSE as below:

$$PSNR = 10 \log_{10} \frac{a_{max}^2}{MSE} \qquad\qquad 2.4$$

$a_{max}$ is 255 when images are represented using 8 bits per sample. PSNR is widely used for evaluation and comparison between different video codec. Huynh-Thu and Ghanbari [] demonstrated that PSNR is reliable for quantify visual image quality when the video content and codec type keep the same. Moreover, as long as the content is changed, PSNR could not work well for assessing the video quality. Between the loss image and video compression, the PSNR values are typically between 30dB and 50dB, where the higher one is better. When the MSE score of two images is equal to zero, the PSNR of them will approach to infinite.

These two main traditional objective methods have marked advantages: easy to calculate and strong theoretical meaning. But they only use mathematical statistics on the difference of pixels, with discarding the correlation between pixels and pixels, also ignore the HVS (Human Visual System) perception. In some situation, these methods cannot accord with the subjective determination from human being. Many researchers exerted their effort on develop these method. [16][20][21][22][23]

### 2.5.3   SSIM (Structural Similarity Index)

The **structural similarity** (SSIM) **index** is a method for measuring the similarity between two images. This method is designed to improve on traditional image quality assessments. With considering that HVS (human visual system) is used to get structural information from the objective visual field, the basic thought of SSIM is compare the similarity of two images instead of calculating the difference between them.The graphic below shows the whole SSIM system:

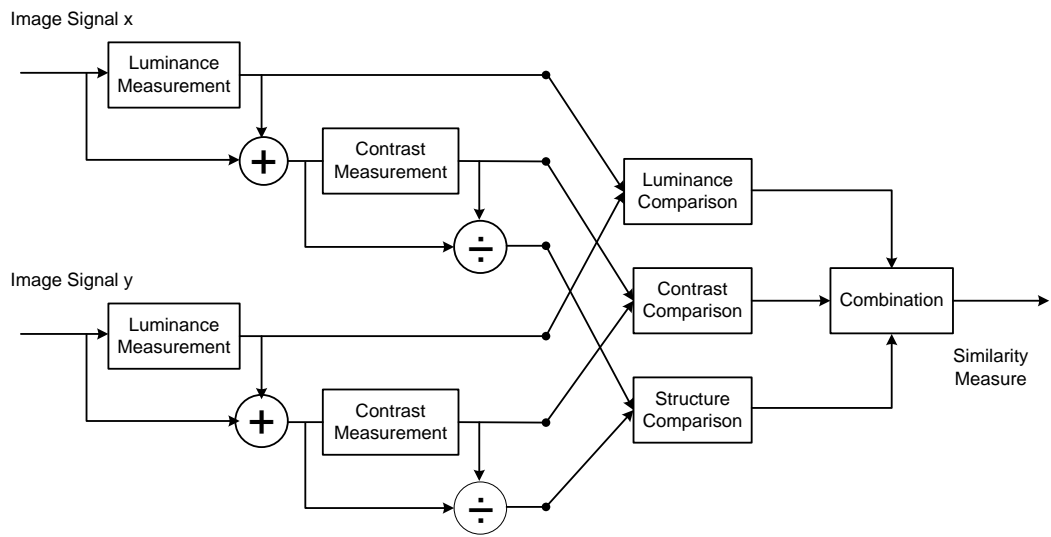

Figure 7   SSIM system

For overall similarity measure, the SSIM system consists of three separates comparisons: luminance, contrast and structure. The objective image will be cut into various windows and calculate to a SSIM metric. The SSIM measure between two image signals x and y of a $N \times N$ window size is defined as below:

$$SSIM(x, y) = [l(x,y)]^{\alpha}[c(x,y)]^{\beta}[s(x,y)]^{\gamma} \qquad 2.5$$

In this formula, $\alpha, \beta, \gamma$ are the weighting coefficients, $\alpha > 0, \beta > 0, \gamma > 0$ ; $l(x,y)$, $c(x,y)$, $s(x,y)$ indicate the function of luminance, contrast, and structure, respectively.

The three comparisons are defined as below:

$$l(x,y) = \frac{2\mu_x\mu_y + C_1}{\mu_x{}^2\mu_y{}^2 + C_1} \quad C_1 = (K_1L)^2 \qquad 2.6$$

$$c(x,y) = \frac{2\sigma_x\sigma_y + C_2}{\sigma_x{}^2\sigma_y{}^2 + C_2} \quad C_2 = (K_2L)^2 \qquad 2.7$$

$$s(x,y) = \frac{2\sigma_{xy} + C_3}{\sigma_x\sigma_y + C_3} \qquad 2.8$$

$$\sigma_{xy} = \frac{1}{N-1}\sum_{i=1}^{N}(x_i - \mu_x)(y_i - \mu_y) \qquad 2.9$$

where $\mu_x$ is the average of x; $\mu_y$ is the average of y; $\sigma_x$ is the variance of x; $\sigma_y$ is the variance of y; $\sigma_{xy}$ is the covariance of x and y; L represent the dynamic range of pixel value; $K_1$ and $K_2$ is constant , and their default value are 0.01 and 0.03.

By this way, a mean SSIM score can be calculated as the final value from the SSIM metric.

According to Wang's [Image Quality Assessment: From Error Visibility to

Structural Similarity] experiment outcome, better evaluation result could be obtained by SSIM system compare to PSNR and MSE. At the same time, SSIM still could not completely solve the problem that caused in PSNR and MSE measurement. Two images which are judged to be with different visual image quality through subjective evaluation still may get the same SSIM score.

In this experiment, SSIM and MSE are both used for evaluation. The parameter fixed in SSIM measuring will be introduced in Chapter.

# Chapter 3  Design and Simulation

### 3.1 Simulation Frame

In the last chapter, some background theory of this project was reviewed. To repairing low quality video enhancement algorithm, we need to simulate the off line process and run time filtering process.

A step by step procedure that common for each embodiment is described in the following figure:

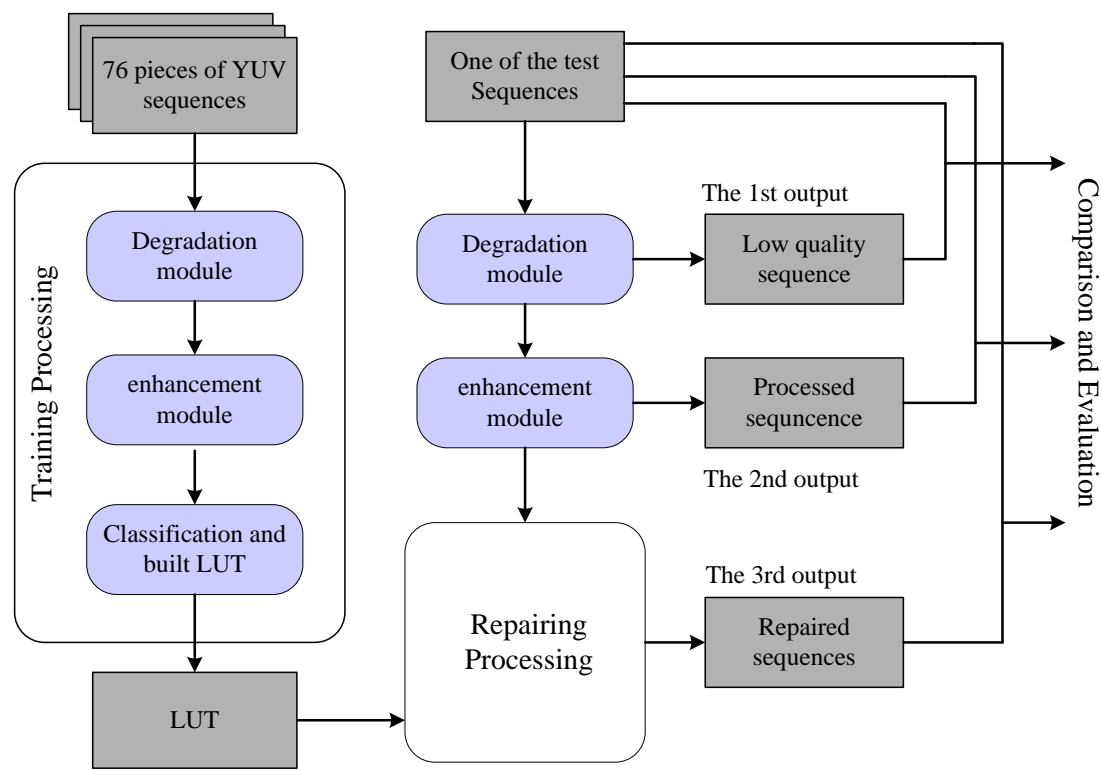

Figure 8   The common procedure of the three embodiments

For experiment, each video process modules dispatched in the above flow-process diagram will be carried out by C programming based on its relevant algorithm.

(a) The degradation of the test sequences will be treated as the low quality videos which need to be enhanced.

(b) In order to enable an MSE and SSIM calculation, the original high quality test sequences will be compared with the result sequences for reference.

(c)Plenty of YUV materials should be trained to get optimized coefficients in the each training process that is individual for relevant embodiment. In this experiment, 76 YUV sequences in different format are used.

(d) The degradation module and the enhancement module in the training process are the same as the one used on the test sequence.

The following sections will describe the simulation in details such as the processing algorithm selected, classification method used, and the coefficients optimization.

## 3.2 Processing Module Selecting

For the three embodiment of this project, basic algorithms will be chosen for clear experiment results. Beside this, these algorithms which usually are simple will not need too much time to compute and make the experiment easier to achieve. Therefore, the selection details are determined as following:

### 3.2.1 Repairing coding artifacts reduction algorithm

The compression module is simulated as degrading process by programming in C language based on JPEG compression standard (http://www.ijg.org/). Here, the JPEG quality is finally set to be 20 to ensure sufficient coding artifacts for sharp contrast with the other output. However, the details of the sequences should not be completely removed in this quality level. The screen shot of "bord" YUV sequences processed after compression process is represented as below:

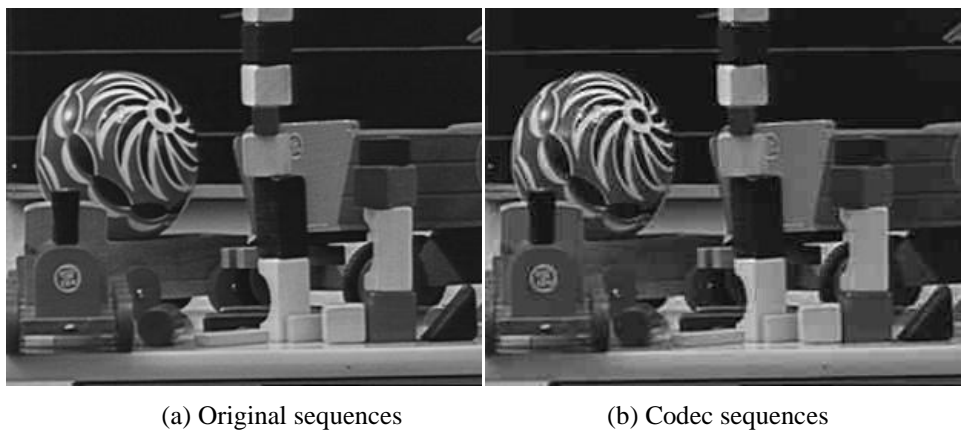

(a) Original sequences          (b) Codec sequences

Figure 9   The effect of Compression

A program of Gaussian filter is used to simulate a low-quality coding artifact reduction algorithm. It can reduce higher frequency but still result in lacking of fine details. In order to obtain suitable effect, the radius parameter is finally fixed to be 2 pixels. This parameter specified the radius of the Gaussian, without counting the center pixel.

Hence, the Gaussian block is a 5-pixel aperture. Sigma express the standard deviation of the Gaussian, in pixels, is eventually set to be 1. The screen shot of "bord" YUV sequences processed after Gaussian filter is represented in the next page:

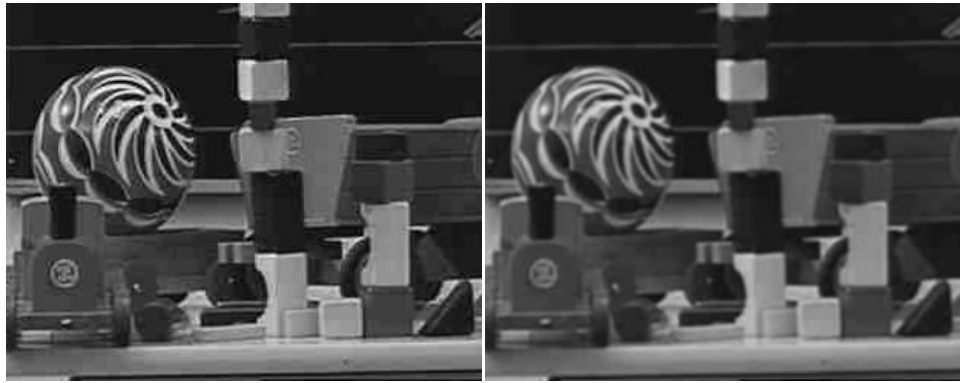

(a) Codec sequence          (b) Codex sequence after Gaussian blurring

Figure 10 The effect after Gaussian blurring

### 3.2.1 Repairing deblurring algorithm

In repairing deblurring algorithm embodiment, Gaussian filter program changed to be the degradation module. As the characteristic briefed in last section, the radius parameter is still fixed to be 2 pixels, and the sigma parameter express is set to be 1 because the visible effect of this process is enough for observation. The screen shot of "bord" YUV sequences processed after compression process is represented as below:

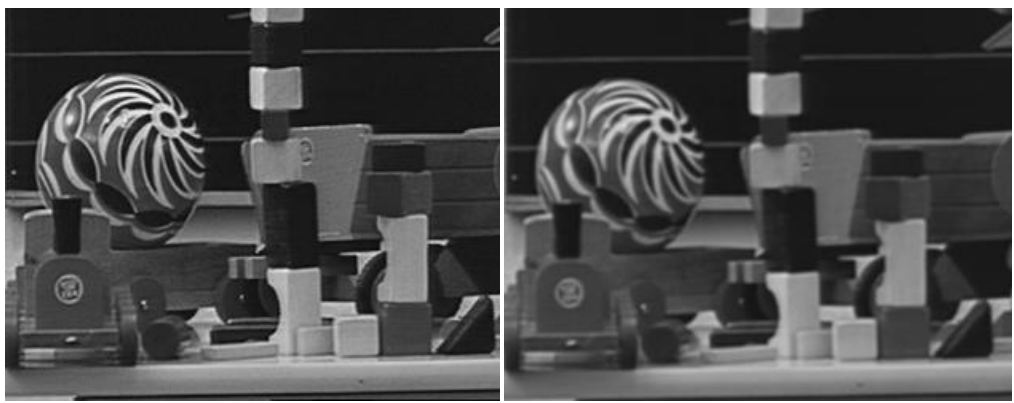

(a) Original sequence          (b) Blurred sequence

Figure 11 Gaussian Blur

A peaking filter program simulated the video enhancement module for de-blurring. In this program, alpha is fixed to be 0.2, and three coefficients are fixed as (−alpha; 1+2×alpha; −alpha) for three adjacent pixels' value. So the peaking filtering process is implemented with the coefficients (−0.2; 1.4; −0.2) in both the horizontal and vertical directions. The effect of this process will be shown in the next figure.

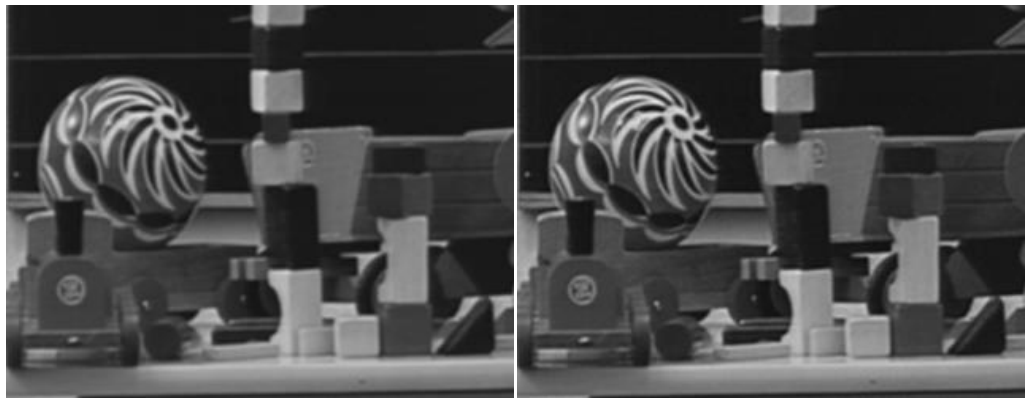

(a) Blurred sequence              (b) After peaking filter

Figure 12 Blurred sequence after peaking filter

### 3.2.1 Repairing deblurring algorithm

A Down sampling process was simulated by programming to get a low resolution sequence from the original one. The output sequence is a quarter of the test sequence. In this program, the original was cut into 4-pixel block, then, the pixel value is the average of the corresponding block in test sequence.

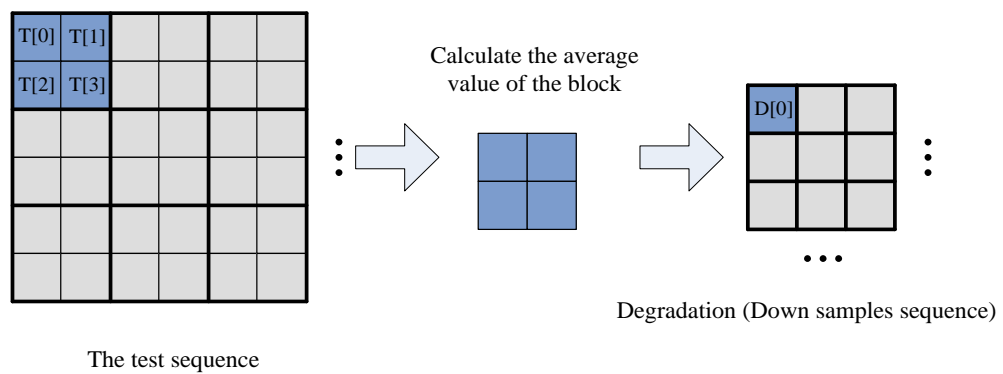

Figure 13 Down sampling process

This graphic is describing the correspondence between the test sequence and the degradation which is the expected low quality sequence. T [0-3] is a 4-pixel block in the test sequence, while D [0] is one pixel in the Degradation.

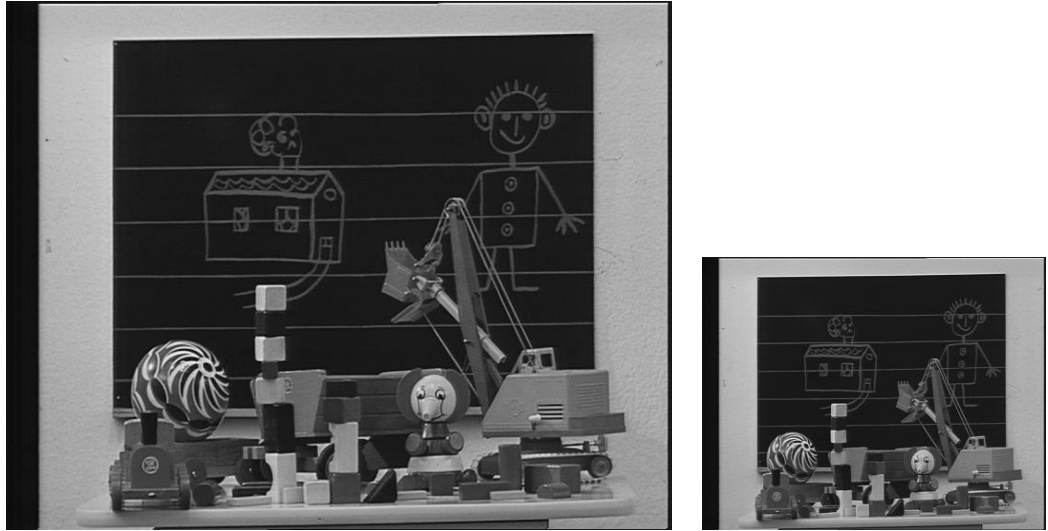

(a) Original sequence                    (b) 1:4 Down sampled sequence

Figure 14 Down sampling process

A bilinear image processing program is simulated to be the resolution up scaling program. This time, the low resolution sequence is the input, and it is still cut into blocks by 4-pixel. A new 4-pixel block can be calculated by the following formula:

$$B[0] = (0.75 \times D[0] + 0.25 \times D[1]) \times 0.75 + (0.75 \times D[2] + 0.25 \times D[3]) \times 0.25$$

$$B[1] = (0.25 \times D[0] + 0.75 \times D[1]) \times 0.75 + (0.25 \times D[2] + 0.75 \times D[3]) \times 0.25$$

$$B[2] = (0.75 \times D[0] + 0.25 \times D[1]) \times 0.25 + (0.75 \times D[2] + 0.25 \times D[3]) \times 0.75$$

$$B[3] = (0.25 \times D[0] + 0.75 \times D[1]) \times 0.25 + (0.25 \times D[2] + 0.75 \times D[3]) \times 0.75$$

Here, D [0-3] expressed the pixel block in the input down sampled sequence, B [0-3] expressed the pixel block in the desired sequence. The relation of these two sequences is described as below:

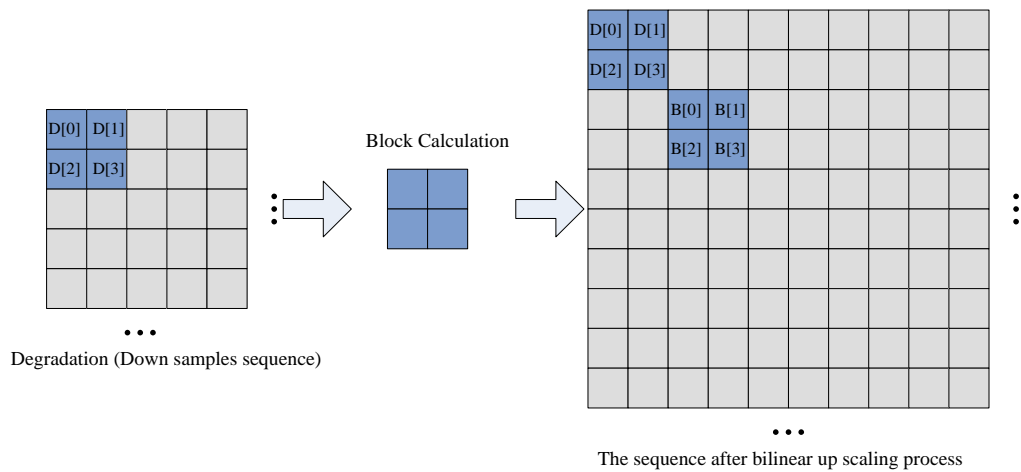

Figure 15 Down sampling process

The effect of this process is depicted as below:

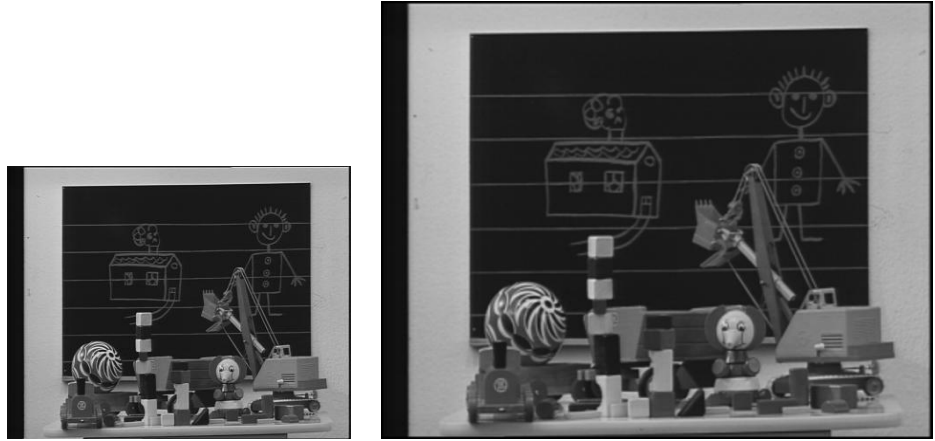

(a) Down-scaled sequence                    (b) After up scaling

Figure 16 Bilinear up scaling process

## 3.3 The classification method used in this project

A diamond shaped aperture including 13 pixels will be employed as a filter aperture in classification during both training and filtering process. This is also used in repairing procedure (the last procedure in filtering process). After this, the central pixel in each pixel's relevant aperture will be the output pixel. The conversion of the data based on the filter aperture is despite detailed in figure 8 as below:

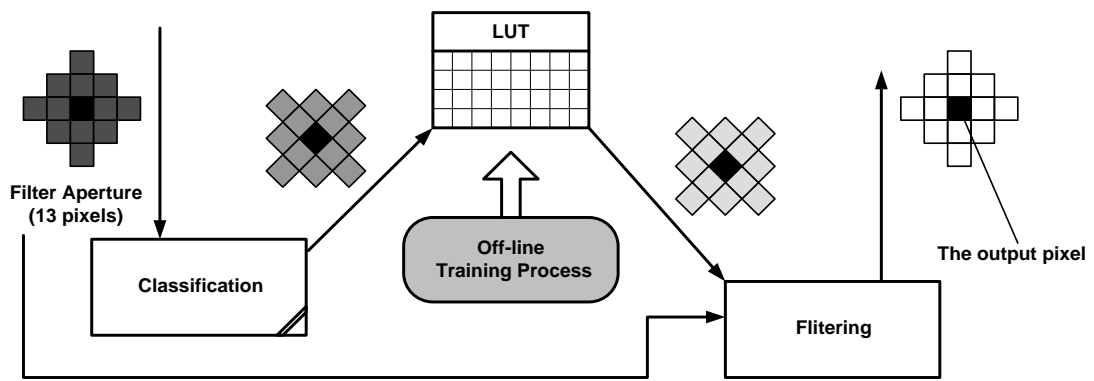

(Classification method used here is ADRC plus one of the complexity determined method)

Figure 17 The diagram of the filter aperture used in filtering process

In this project, different classification method will be tailored for the three embodiments. After consulted the paper [1] [7] [9] [14] [15-18], the appropriate classification methods for each embodiment are finally specified to be: ADRC plus STD for repairing coding artifact reduction and deblurring algorithm; ADRC without other classification method for repairing resolution up-scaling algorithm.

### 3.4 Coefficient Optimization

In this low-quality video enhancement modules repairing algorithm, we use two categories data to find out the optimized coefficients. One of them is from the corresponding pixels in the target image, the other category are apertures of the degraded images (the pixels and their neighbor pixels in specific class). They are all classified for a particular class. The LMS Optimization for training process will be described as below:

Firstly, if we suppose $F_{D,c}(i,j)$ is the apertures of the degraded images and $F_{R,c}(j)$ be the corresponding pixels in the target images for a specific class. Then $F_{F,c}(j)$ which are supposed to be the filtered pixels will be obtained the optimal coefficients $w_c(i), i\epsilon[1\ldots n]$ as follows [1]:

$$F_{F,c}(j) = \sum_{i=0}^{n} w_c(i)F_{D,c}(i,j) \qquad 3.1$$

In this equation, j represents a specific aperture belonging to class c; i express the index of the pixels in a aperture and n is the number of pixels in the aperture.

The equation of square error between the target pixels $F_{R,c}(j)$ and the filtered pixels $F_{D,c}(i,j)$ can be expressed as below [1]:

$$e^2 = \sum_{j=1}^{Nc}\big(F_{R,c}(j) - F_{F,c}(j)\big)^2 \qquad 3.2$$

$$e^2 = \sum_{j=1}^{Nc}\big[F_{R,c}(j) - \sum_{j=1}^{Nc} w_c(i)F_{D,c}(i,j)\big]^2 \qquad 3.3$$

Where $N_c$ indicates the number of training samples belonging to a particular class c. the first derivative of $e^2$ to $w_c(k)$, $k\epsilon[1\ldots n]$ is supposed to be zero to minimize $e^2$.

$$\frac{\partial c^2}{\partial W_{c(k)}} = \sum_{j=1}^{Nc} 2F_{R,c}(j) \times \big[F_{R,c}(j) - \sum_{j=1}^{Nc} w_c(i)F_{D,c}(i,j)\big] = 0 \qquad 3.4$$

Since there is no constant in this equation, we can easily solve this problem by using Gaussian elimination which is mentioned before, we will get the optimal coefficients as following [1]:

$$\begin{bmatrix} w_c(1) \\ w_c(2) \\ \ldots \\ w_c(n) \end{bmatrix} = \begin{bmatrix} \sum_{j=1}^{Nc} F_{D,c}(1,j)F_{D,c}(1,j) & \ldots & \sum_{j=1}^{Nc} F_{D,c}(1,j)F_{D,c}(n,j) \\ \sum_{j=1}^{Nc} F_{D,c}(2,j)F_{D,c}(1,j) & \ldots & \sum_{j=1}^{Nc} F_{D,c}(2,j)F_{D,c}(n,j) \\ \ldots & \ldots & \ldots \\ \sum_{j=1}^{Nc} F_{D,c}(n,j)F_{D,c}(1,j) & \ldots & \sum_{j=1}^{Nc} F_{D,c}(n,j)F_{D,c}(n,j) \end{bmatrix}^{-1} \begin{bmatrix} \sum_{j=1}^{Nc} F_{D,c}(1,j)F_{R,c}(j) \\ \sum_{j=1}^{Nc} F_{D,c}(1,j)F_{R,c}(j) \\ \ldots \\ \sum_{j=1}^{Nc} F_{D,c}(1,j)F_{R,c}(j) \end{bmatrix} \qquad 3.5$$

To be mentioned, aim to get well-performed results from repairing, a large amount of training samples are preferred to get comparative precision coefficients.

# Chapter 4  Experiments and Results

The experiment consists of three embodiments including de-blocking, de-blurring, and up-scaling to estimate the trained filter algorithm for repairing low-quality video processing modules. The performance of every process for three embodiments will all be evaluated by SSIM and MSE measurement. The experiment results and analysis of each algorithm are indicated respectively.

## 4.1 Test sequences

To evaluate the described LSM trained filter algorithm, 9 video sequences with comparatively more details is selected in order to get a flagrant contrast in each video enhancement process. The graphics below are showing the screen shots of the selected test sequences.

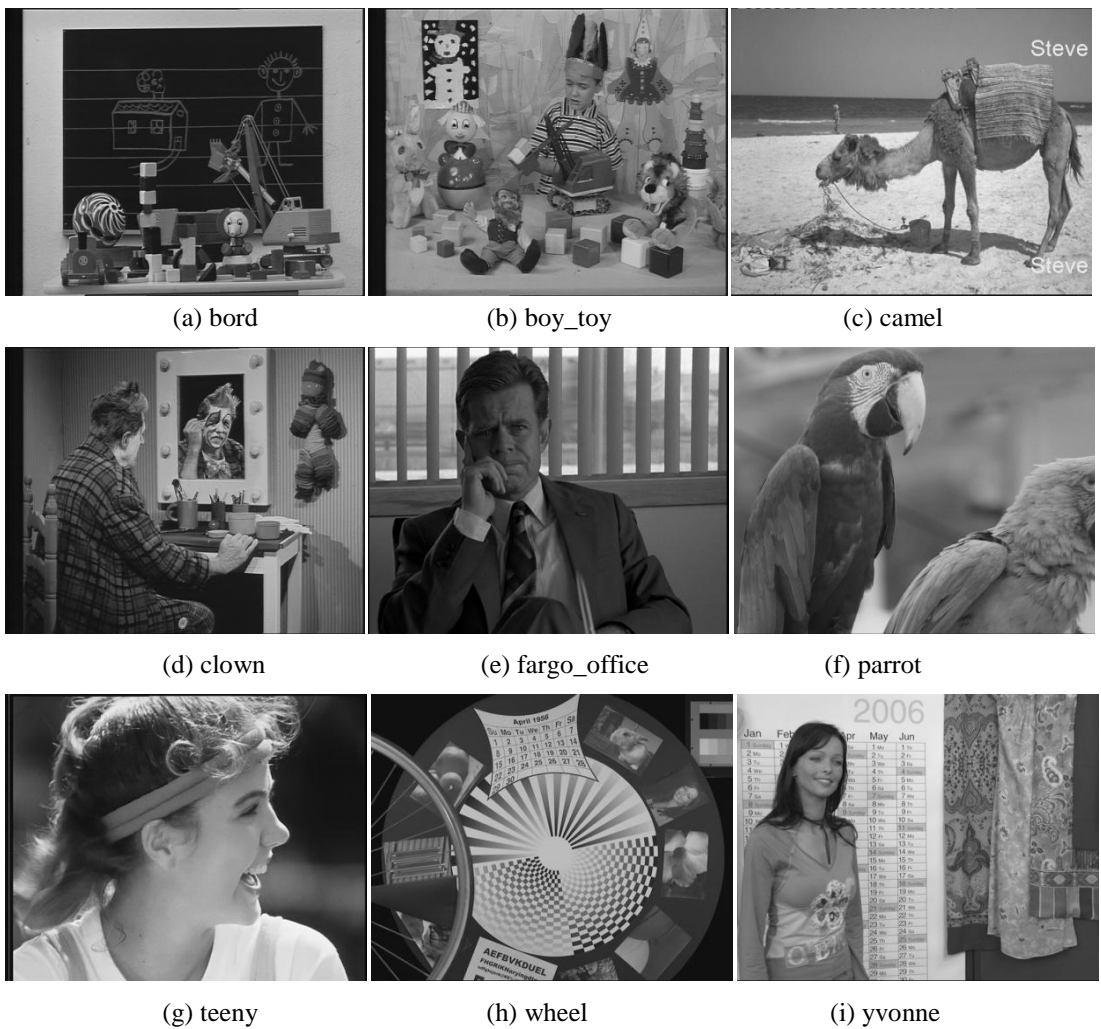

(a) bord                 (b) boy_toy                 (c) camel

(d) clown             (e) fargo_office             (f) parrot

(g) teeny                 (h) wheel                 (i) yvonne

Figure 18 Screen shots of the test sequences

The test sequences used here are all YUV format which is a kind of color space codec format for image and video. This format is invented for display color sequences in black and white infrastructure. Also, It has another advantage is that it can save bandwidth, comparing with RGB format.

The YUV model is made up of three components: Y component contains luminance (or luma) information which display black and white signal, while U and V components represent chrominance or chrome. Only Y component will be processed in this experiment.

YUV format have several different forms decided by the way that luminance signals and chrominance/chrome signals be sampled. YUV format can represent videos in forms such as YUV 4:2:2, YUV 4:4:4, YUV 4:2:0 and YUV 4:1:1. YUV4:2:2 form is the most widely used, and YUV 4:4:4 is only used in high-end video processing equipments.

In this experiment, the selected sequences are all in 4:2:2 form with the same size of 720×576. Every luminance value is represented in 8-bit. Y component will be extracted for processing at the beginning of each process. The property of the test sequences are displaying in the following table:

Table I   THE PROPERTY OF THE TEST SEQUENCES

| Test Sequences | | | | | | |
|---|---|---|---|---|---|---|
| No. | Names | Format | Image frequency | Frames | Width | Height |
| 1 | bord.yuv | YUV422-8bit | 50HZ | 1 | 720 | 576 |
| 2 | boy_toy.yuv | YUV422-8bit | 50HZ | 1 | 720 | 576 |
| 3 | camel.yuv | YUV422-8bit | 24HZ | 47 | 720 | 576 |
| 4 | clown.yuv | YUV422-8bit | 50HZ | 1 | 720 | 576 |
| 5 | parrot.yuv | YUV422-8bit | 50HZ | 98 | 720 | 576 |
| 6 | teeny.yuv | YUV422-8bit | 50HZ | 36 | 720 | 576 |
| 7 | wheel.yuv | YUV422-8bit | 25HZ | 40 | 720 | 576 |
| 8 | yvonne.yuv | YUV422-8bit | 25HZ | 50 | 720 | 576 |
| 9 | fargo_office.yuv | YUV422-8bit | 50HZ | 107 | 720 | 576 |

## 4.2 Repair for the Coding artifacts reduction algorithm

The training process for repairing coding artifacts reduction was implemented in the following figures:

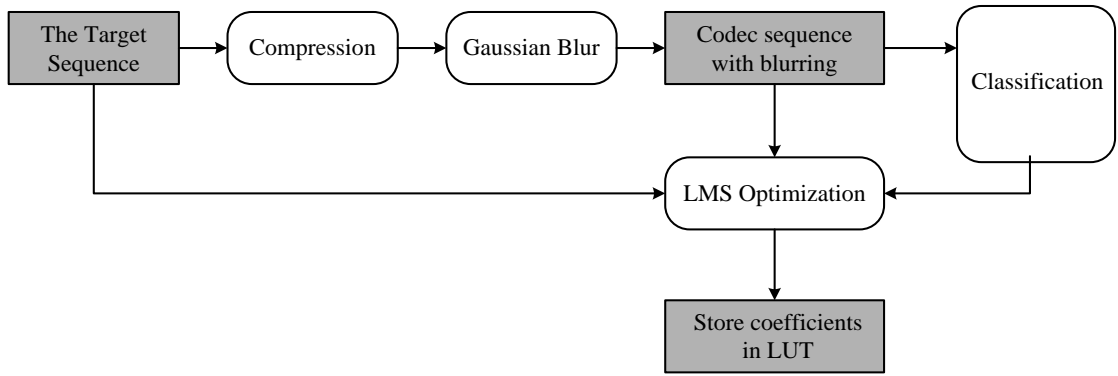

Figure 19  Training process for repairing de-blocking algorithm

The target image mentioned in the above figure is the original test YUV sequences. The output of codec module, Gaussian blurred module and repairing process will be recorded for comparing with it.

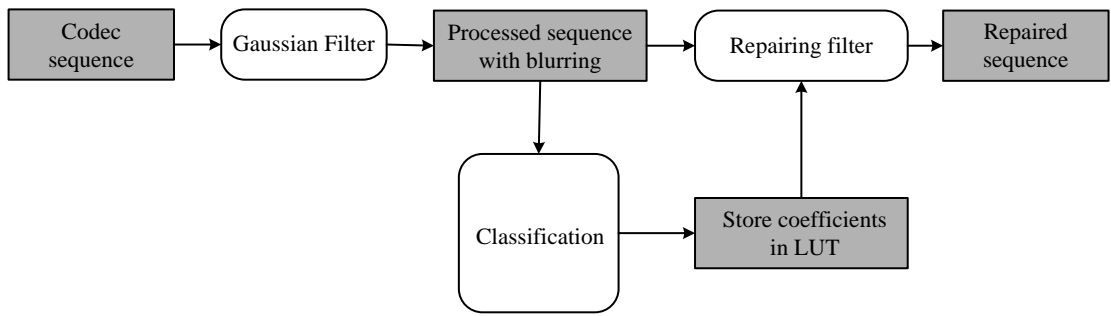

Figure 20  Filtering process for repairing de-blocking algorithm

In repairing process, the classification method is selected to be the combination of ADRC and standard deviation as we mentioned before. After implement the training process program, an output data named "_table_B" obtain the desired coefficients and the blurred codec sequence will be used as the input in the repairing process program.

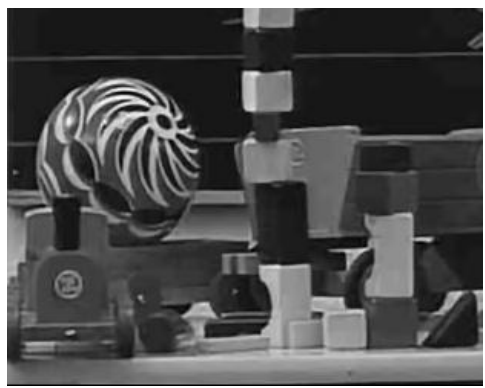

Figure 21  Result sequence of Repairing for de-blocking algorithm

The following graphics represent an entire comparison of the three output sequences from the whole experiment process of the "bord.yuv".

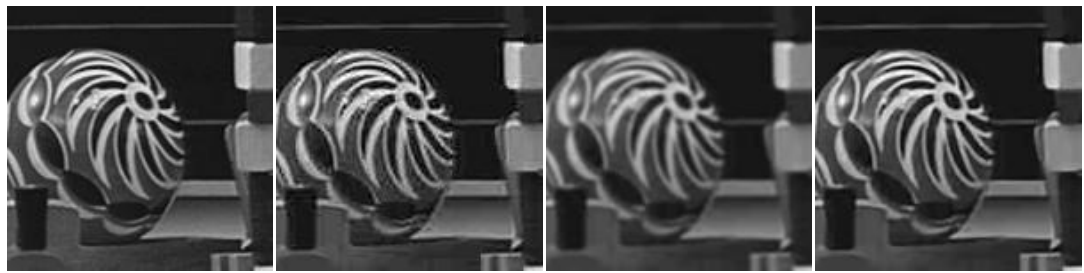

(a) Original sequences    (b) Codec sequences    (c) "b" after blurring    (b) Repaired sequences

Figure 22  Output sequences in repairing de-blocking algorithm

From the graphics above, we can see that the video quality has been significantly increased by the proposed method. It deducted the coding artifacts in picture (b), and also protects the image details compared with the picture (c). However, it is still not as good as the target sequence showed in picture (a). The snapshot of other test sequences will be shown in the appendix. The table below is the MES score and SSIM score calculated from the experiment results. As same as the visible effect of above graphics, the SSIM score is more proximal to 1 with the processes operate step by step. To be mentioned, as we can see in the MSE column, the MSE score of the sequences after Gaussian Blur is conspicuously higher than others. This may because MSE score is calculate pixel by pixel while SSIM score is computed widow by window, then, the Gaussian blur which is operate in a 5-pixel aperture lead to be far different from the original sequence when estimate on each pixels.

Table II    THE MSE SCORE AND SSIM SCORE OF REPAIRING CODING ARTIFACT REDUCTION

| Test Sequence | | Repairing Coding Artifact Reduction | | | | | |
|---|---|---|---|---|---|---|---|
| | | MSE | | | SSIM | | |
| No. | Names | Blocky | Gaussian Blur | Repair | Blocky | Gaussian Blur | Repair |
| 1 | bord | 27.61 | 57.04 | 20.98 | 0.8475 | 0.9095 | 0.9232 |
| 2 | boy_toy | 37.97 | 61.88 | 28.21 | 0.8422 | 0.8998 | 0.9054 |
| 3 | camel | 40.03 | 56.52 | 31.93 | 0.9004 | 0.9086 | 0.9157 |
| 4 | clown | 32.49 | 46.02 | 26.11 | 0.8748 | 0.9037 | 0.9173 |
| 5 | parrot | 21.03 | 20.75 | 17.85 | 0.8897 | 0.8965 | 0.8993 |
| 6 | teeny | 23.22 | 31.57 | 18.4 | 0.9239 | 0.9289 | 0.9308 |
| 7 | wheel | 51.11 | 87.19 | 30.5 | 0.9375 | 0.9315 | 0.9449 |
| 8 | yvonne | 45.51 | 69.09 | 35.28 | 0.9004 | 0.9113 | 0.9222 |
| 9 | fargo_office | 19.27 | 27.75 | 15.97 | 0.9313 | 0.9495 | 0.9621 |

### 4.3 Repair for the De-blurring algorithm

The training process for repairing coding artifacts reduction is described in the below figures:

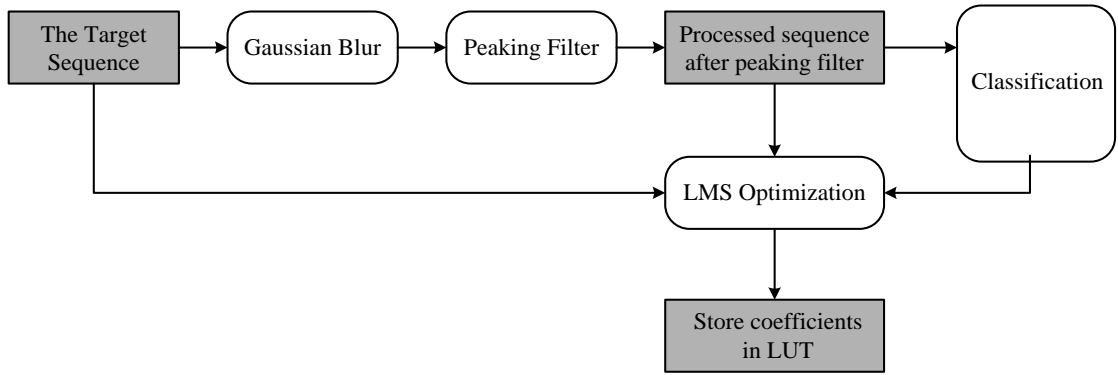

Figure 23  Training process for repairing deblurring algorithm

The filtering process for repairing deblurring algorithm is depicted as below:

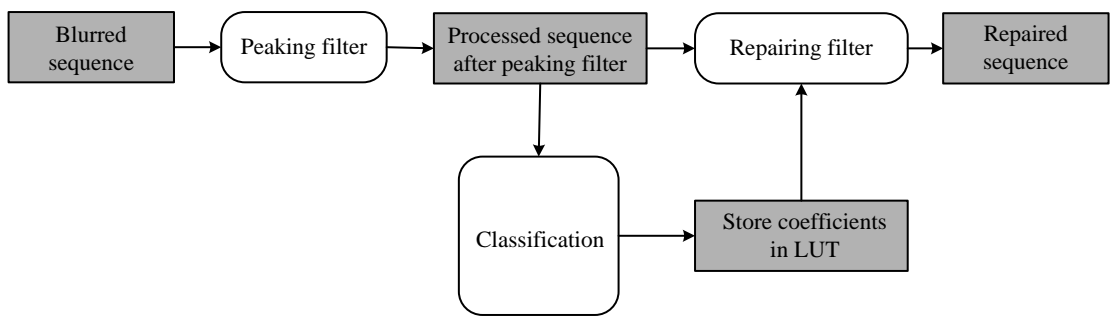

Figure 24  Filtering process for repairing deblurring algorithm

As same as repairing coding artifact reduction, the classification method used here is also the combination of ADRC and standard deviation. The training process which embed the blurring process and peaking filter process carry out the desired coefficients in LUT for deblurring-repaired process. A picture of the screenshot of the output sequence is shown:

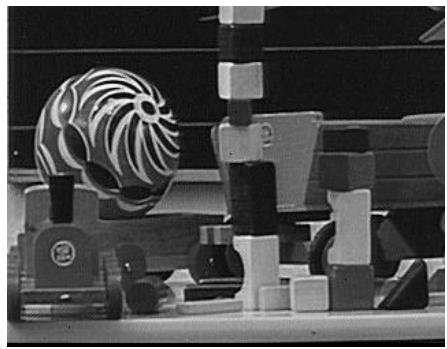

Figure 25  Result sequence of Repairing for deblurring algorithm

A whole comparison of three output sequences from the whole experiment process of the "bord.yuv" is depicted in the following.

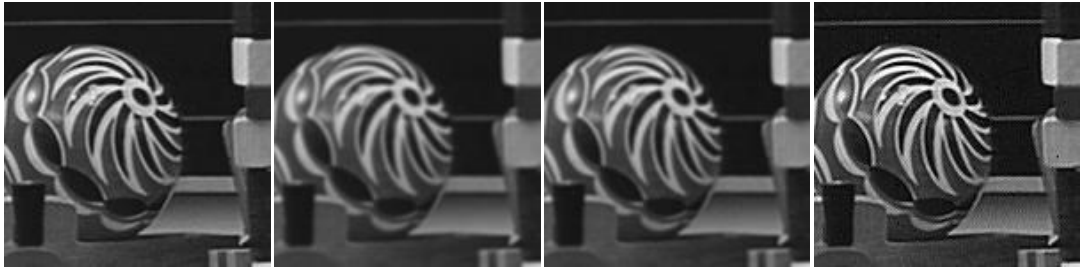

(a) Original sequences    (b) blurred sequences    (c) "(b)" after peaking filter (d) Repaired sequences

Figure 26 Output sequences in repairing deblurring algorithm

As we can see from the graphics above, the result sequence is sharp, and has legible details. It is nice repaired the sequence "c", and visible almost the same as the target sequence (a). The snapshot of other test sequences will be shown in the appendix. The MES score and SSIM score of the output sequences in this embodiment are calculated as below. The SSIM score are stable increased and on the verge of 1, while the MSE score are steady reduced.

Table III  THE MSE SCORE AND SSIM SCORE OF REPAIRING RESOLUTION UP SCALING

| Test Sequence | | Repairing resolution up scaling | | | | | |
|---|---|---|---|---|---|---|---|
| | | MSE | | | SSIM | | |
| No. | Names | GB | Sharpen | Repair | GB | Sharpen | Repair |
| 1 | bord | 48.32 | 34.34 | 30.09 | 0.9379 | 0.9526 | 0.9649 |
| 2 | boy_toy | 50.05 | 35.15 | 28.5 | 0.9365 | 0.9532 | 0.9573 |
| 3 | camel | 44.79 | 34.16 | 25.27 | 0.9447 | 0.9572 | 0.9613 |
| 4 | clown | 35.28 | 25.87 | 22.66 | 0.9349 | 0.9547 | 0.9608 |
| 5 | parrot | 13.94 | 14.49 | 12.14 | 0.9316 | 0.9387 | 0.9434 |
| 6 | teeny | 23.51 | 18.62 | 15.72 | 0.9584 | 0.9644 | 0.9679 |
| 7 | wheel | 74.67 | 52.74 | 48.77 | 0.9513 | 0.9633 | 0.9724 |
| 8 | yvonne | 56.53 | 41.68 | 37.12 | 0.9281 | 0.9457 | 0.9516 |
| 9 | fargo_office | 20.69 | 17.76 | 16.85 | 0.9735 | 0.9757 | 0.9788 |

## 4.4 Repair for the resolution up-scaling algorithm

The training process and repairing process for repairing coding artifacts reduction are described in the following figures:

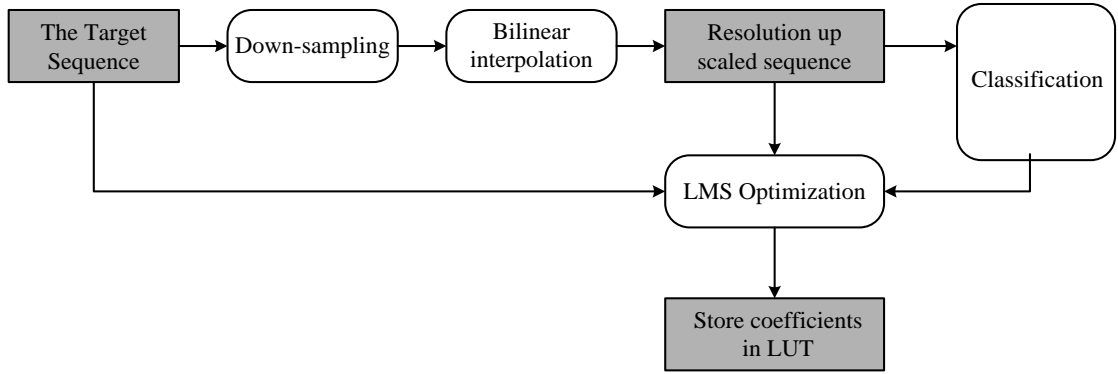

Figure 27 Training process for repairing resolution up-scaling algorithm

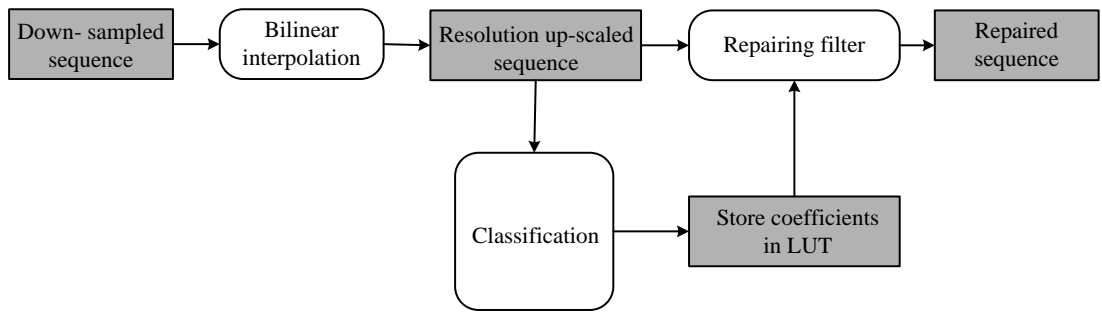

Figure 28 Filtering process for repairing resolution up-scaling algorithm

The training process and repairing process are specified as above. Since "the complexity measures do not contribute to the MSE scores for resolution up-conversion", and "This may be because contrast information is irrelevant for interpolation."[1] In this embodiment, the classification method is simply ADRC. The output sequence of the repairing process is shown as below:

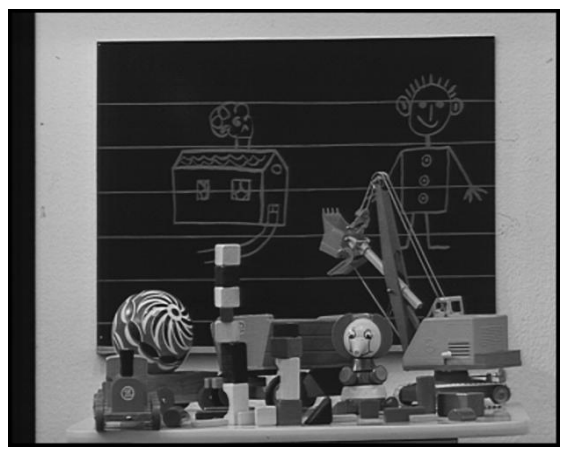

Figure 29 Result sequence for repairing resolution up scaling algorithm

Three output sequences from each process of the "bord.yuv" with comparing to the target sequence are depicted in the following.

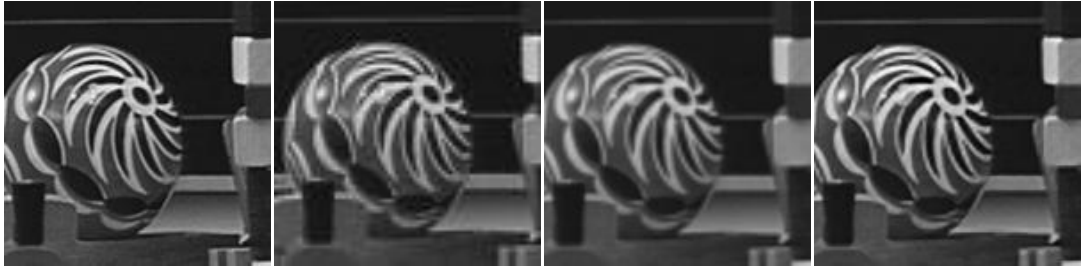

(a) Original sequences (b) Down-sampled sequences (c) "b" after up-scaling (d) Repaired sequences

Figure 30 Output sequences in repairing resolution up scaling algorithm

From the results' comparison, the visible quality of the repaired sequence (d) is highly increased than that of the sequence (c), but it still has a distance from the target sequence (a). A table record the MSE score and SSIM score of the output sequences in this embodiment is below. To be mentioned, the outputs of the down sampling process are small size of the original sequences. Identically, MSE and SSIM score could not be calculated between two images with different size. And this is also not necessary.

Table IV  The MSE score and SSIM score of repairing resolution up scaling algorithm

| Test Sequence | | Repairing resolution up scaling | | | | | |
| --- | --- | --- | --- | --- | --- | --- | --- |
| | | MSE | | | SSIM | | |
| No. | Names | DS | Up scaling | Repair | DS | Up scaling | Repair |
| 1 | bord | | 57.01 | 26.93 | | 0.9206 | 0.9403 |
| 2 | boy_toy | | 57.04 | 25.36 | | 0.9201 | 0.9395 |
| 3 | camel | | 51.36 | 28.17 | | 0.9228 | 0.9381 |
| 4 | clown | | 40.83 | 20.51 | | 0.9184 | 0.9427 |
| 5 | parrot | | 15.28 | 11.85 | | 0.9082 | 0.9099 |
| 6 | teeny | | 27.31 | 13.9 | | 0.9476 | 0.9548 |
| 7 | wheel | | 86.49 | 35.37 | | 0.9342 | 0.959 |
| 8 | yvonne | | 66.67 | 39.58 | | 0.8999 | 0.9226 |
| 9 | fargo_office | | 24.51 | 16.13 | | 0.9618 | 0.9632 |

# Chapter 5   Conclusion

This report attempted to introduce an algorithm of repairing low quality video enhancement module based on trained filter. In this report, the fundamental knowledge was represented firstly. Then, relevant algorithms employed in simulating this algorithm were determined, simulated, and implemented. For a flagrant contrast between before and after in the experiment, 9 sequences were employed as the test sequences. At the end, experiment results were evaluated by MSE and SSIM calculating.

Three embodiments for the basic image enhancement modules, which are coding artifacts reduction, deblurring, resolution up scaling, were carried out in this project. Comparing the results of each process with evaluated by MSE score and SSIM score, the proposed algorithm repaired the low quality sequences well approached to the target sequence. These results evidentially proved the proposed algorithm and showing the nice performance of trained filter.

However, the performance of three embodiments is slightly different due to the different video enhancement module they used. Also, different sequence results in a little difference as well. In the future, if necessary, more experiments can be done to evaluate the application level of the algorithm. More study need to be done for more video enhancement algorithm and the comparisons of repairing them. Moreover, since traditional video enhancement algorithm can be used double or triple time for a better performance, a full contrast can also be done between proposed algorithms and double used traditional algorithm which are usually economical and time-saving in the practical use. From that, a conditional practical use can be estimated.

# APPENDIX I   Diary of Major Milestone Achieved

Plenty of readings including books and papers have been done in this project. Apart from Easter vacation, background learning and research preparing was focused during the second semester. The main methodology of this project has been developed in June. One week was spent on preparing and designs the procedure of the experiment. From then, three embodiments were simulated by programming in C language. After this, test, evaluation and comparison have been done on this experiment. At this semester, most of the time is spent on programming and testing. A table of the major stone achieved is representing the detail as below:

| TASK NAME | DURING DATE |
|---|---|
| Preparing and design the procedure of the experiment | 2010/6/27 |
| Programming and testing for the first embodiment: repairing the coding artifacts reduction algorithm | 2010/7/9 |
| Programming and testing for the second embodiment: repairing the deblurring algorithm | 2010/7/13 |
| Programming and testing for the third embodiment: repairing the | 2010/7/23 |
| Programming for the SSIM and MSE evaluator | 2010/7/30 |
| Experiment and results estimate | 2010/8/13 |
| Finish dissertation and poster | 2010/8/31 |

# APPENDIX II  The snapshot of the outputs from the experiment

1.  The output sequences of repairing for the coding artifacts reduction algorithm

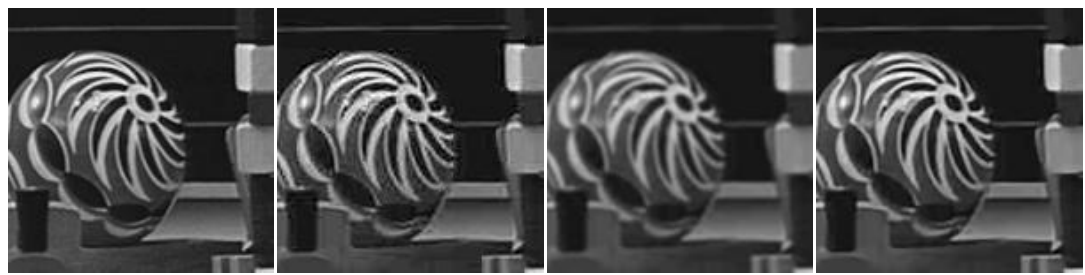

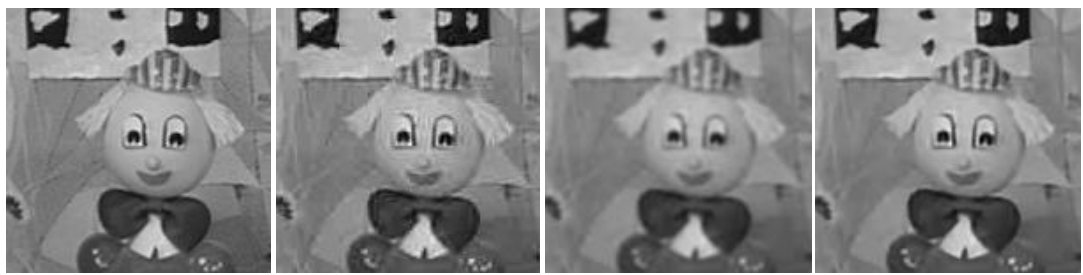

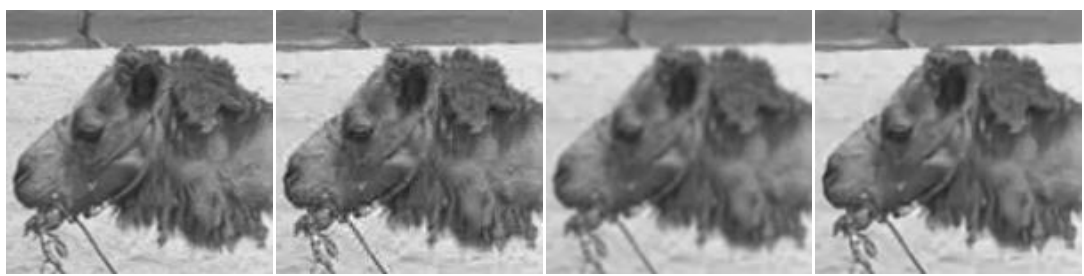

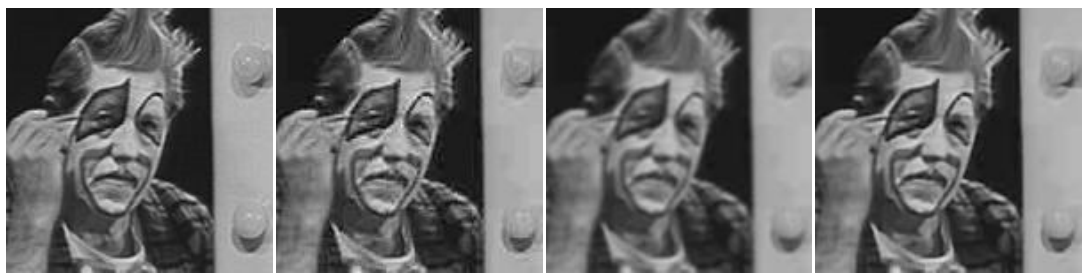

(a) Original sequences     (b) Codec sequences     (c) "b" after blurring     (b) Repaired sequences

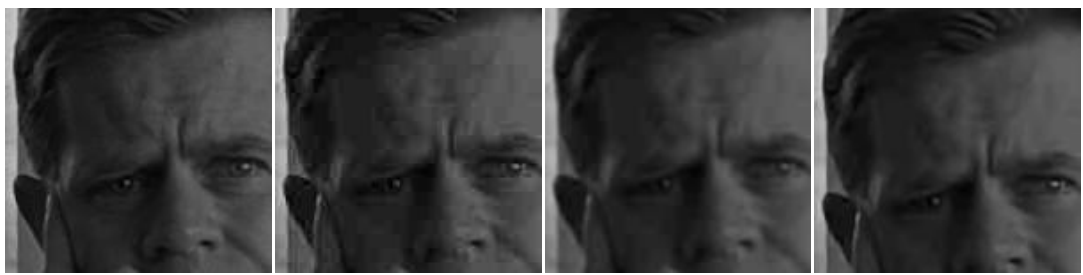

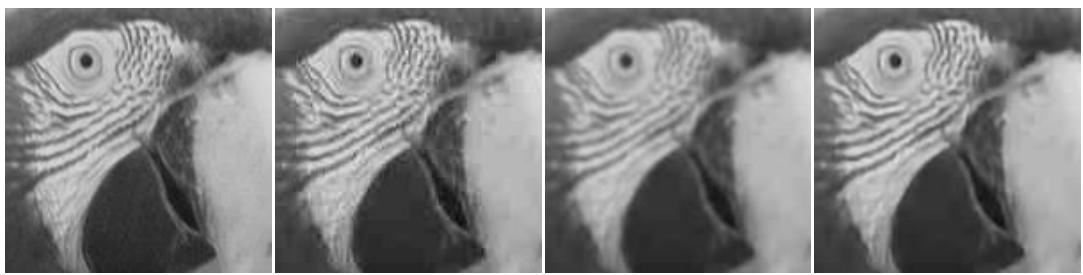

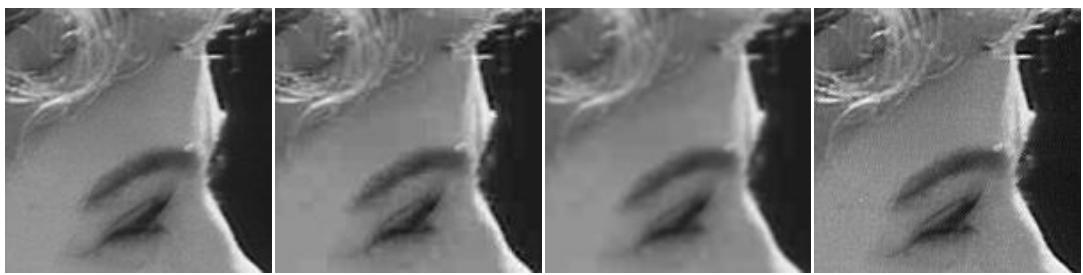

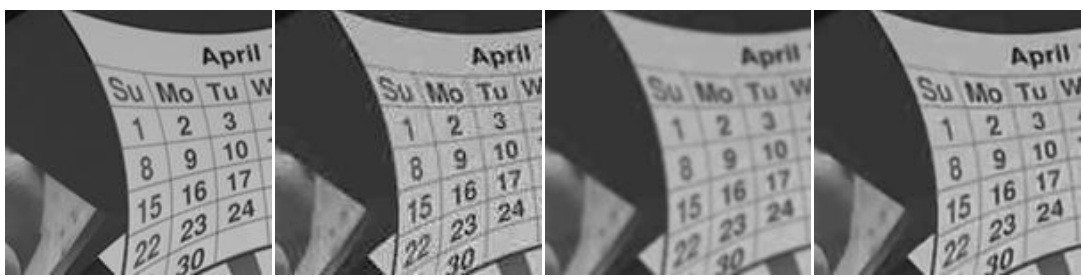

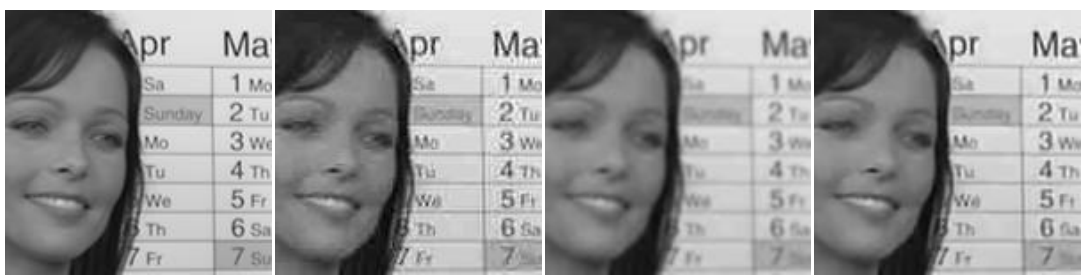

(a) Original sequences     (b) Codec sequences     (c) "b" after blurring     (d) Repaired sequences

2.    The output sequences of repairing for the deblurring algorithm

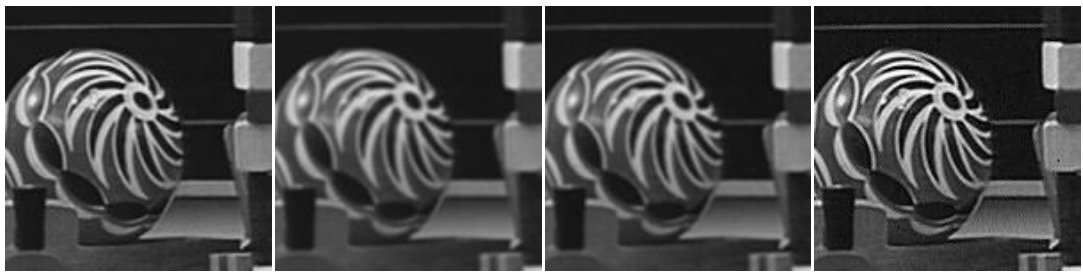

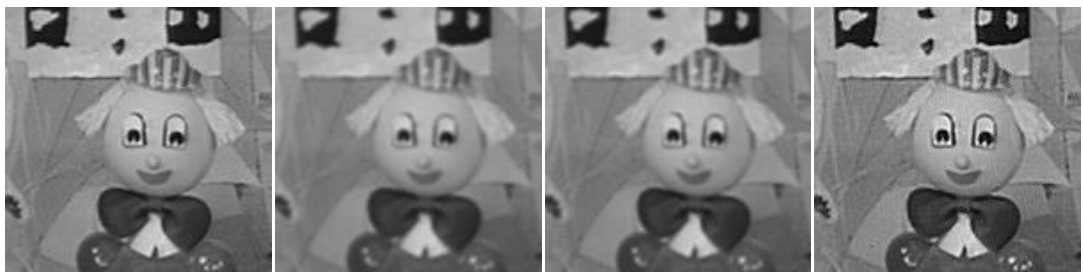

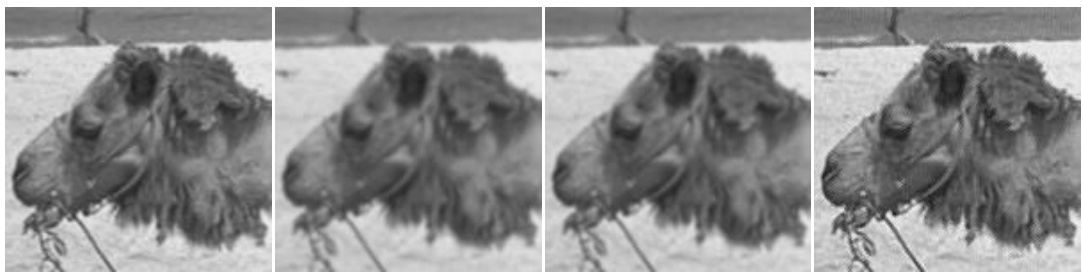

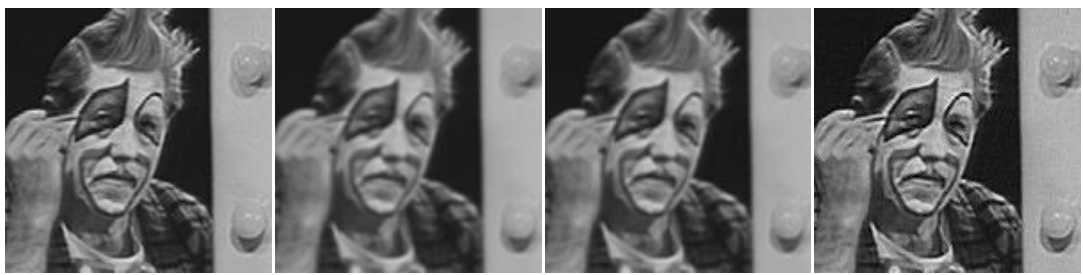

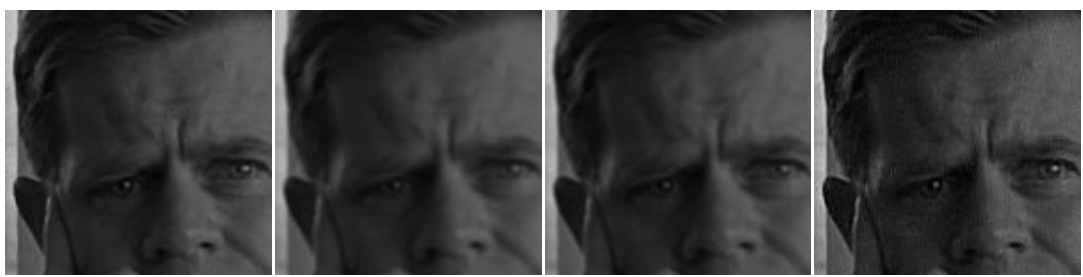

(a) Original sequences    (b) blurred sequences    (c) "b" after peaking filter (d) Repaired sequences

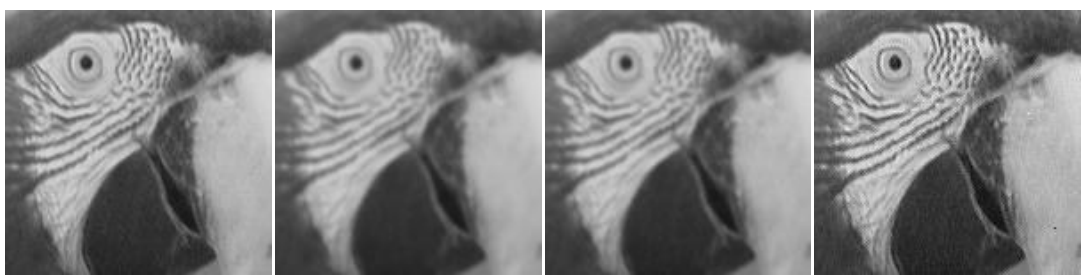

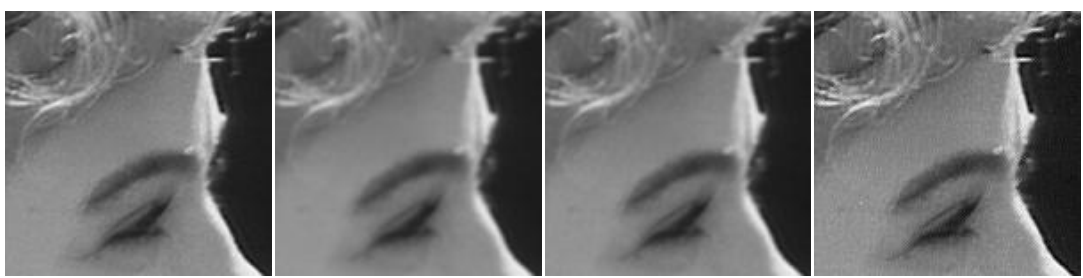

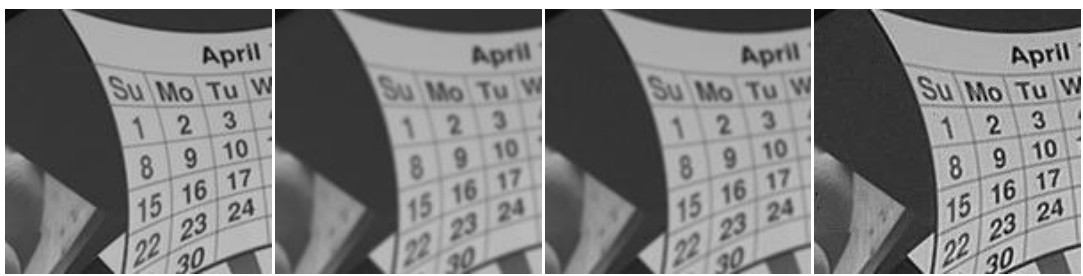

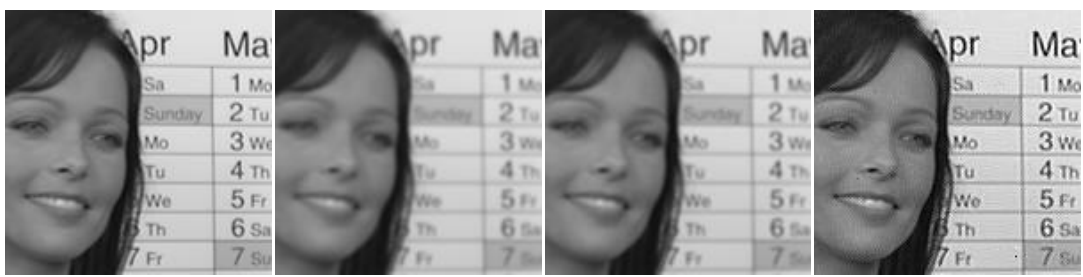

(a) Original sequences    (b) blurred sequences    (c) "b" after peaking filter (d) Repaired sequences

3. The output sequences of repairing for the resolution up scaling algorithm

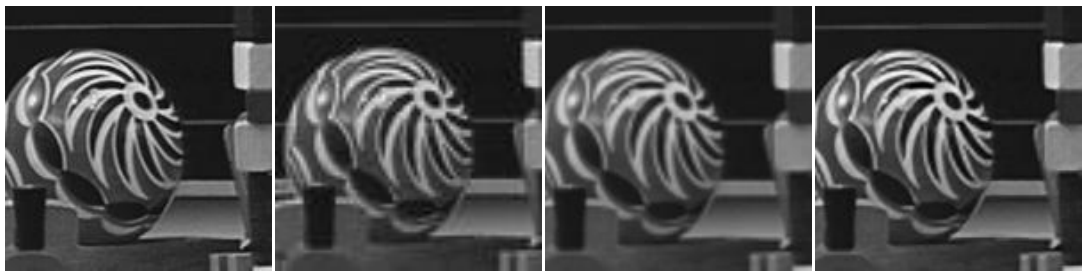

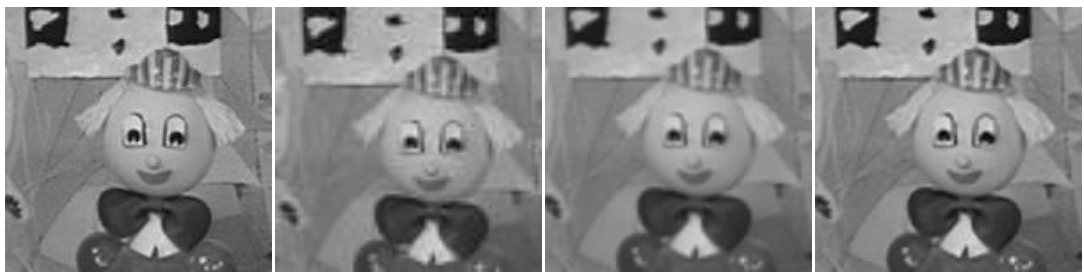

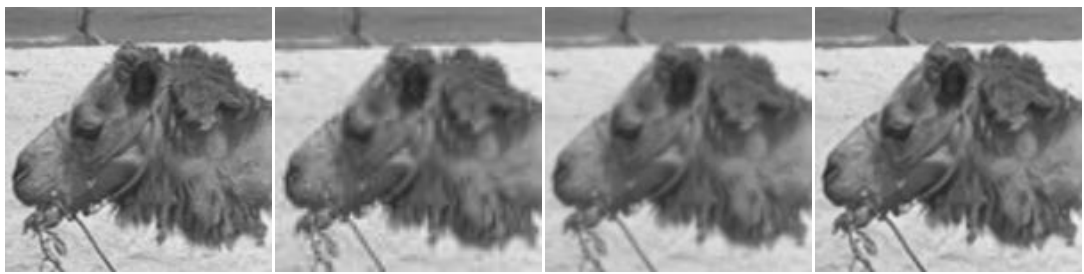

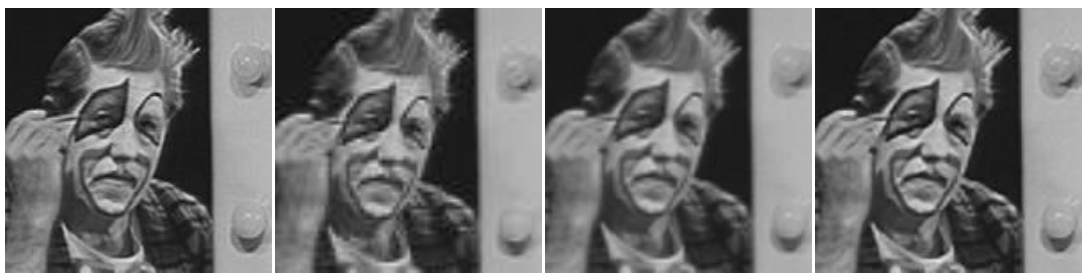

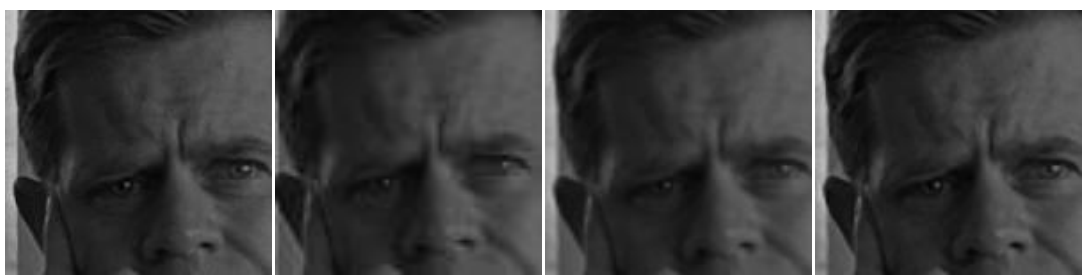

(a) Original sequences (b) Down-sampled sequences (c) "b" after up-scaling (d) Repaired sequences

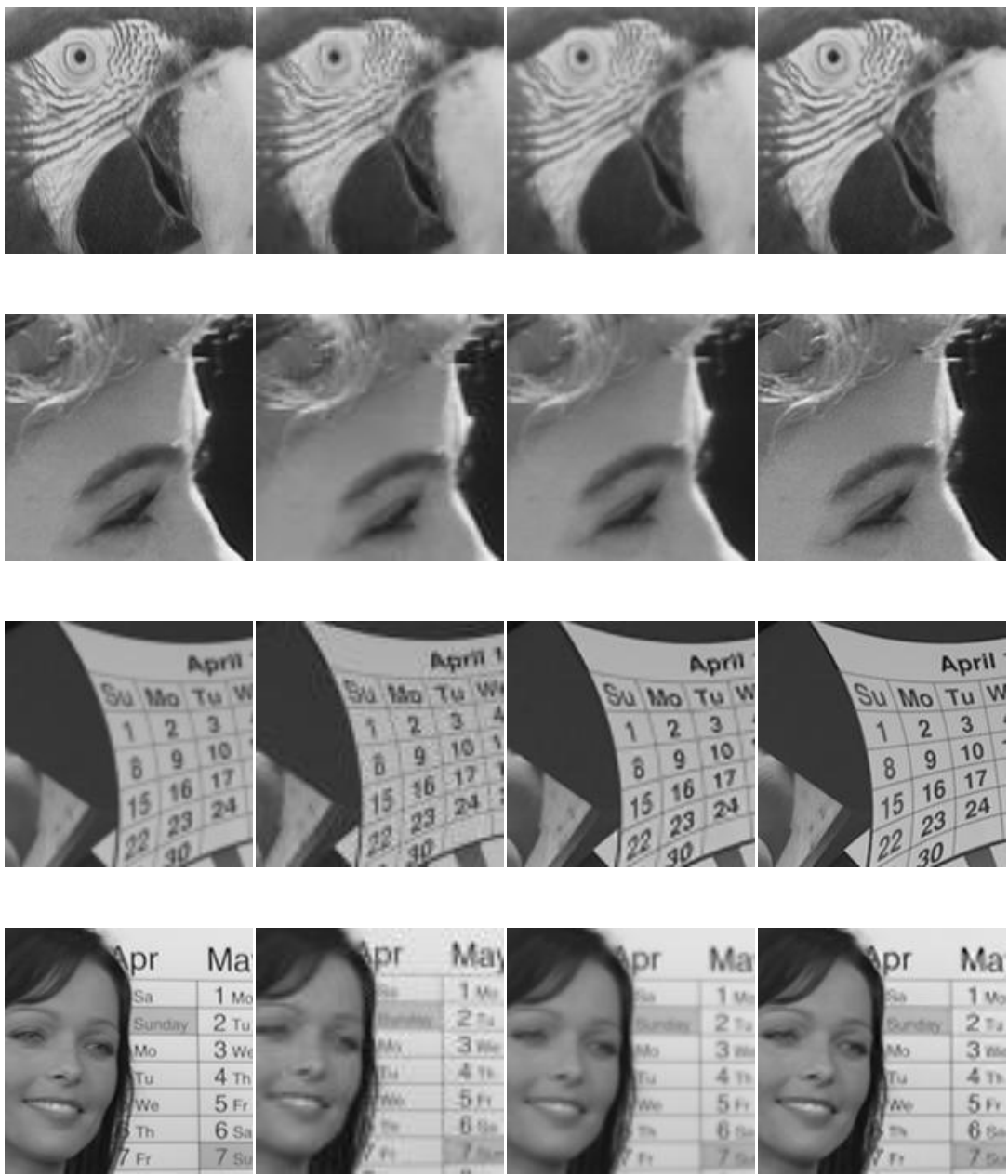

(a) Original sequences (b) Down-sampled sequences (c) "b" after up-scaling (d) Repaired sequences